\RequirePackage{fix-cm}
\documentclass[twocolumn]{svjour3}       
\smartqed  
\usepackage{graphicx}
\usepackage{amsmath}
\usepackage{amssymb}
\usepackage[symbol*]{footmisc}
\usepackage{url}
\usepackage{cite}
\usepackage{algorithm}
\usepackage{algpseudocode}

\usepackage{xcolor}

\DeclareMathOperator{\leaf}{leaf}
\DeclareMathOperator*{\argmin}{arg\,min}

\newcommand\blfootnote[1]{%
  \begingroup
  \renewcommand\thefootnote{}\footnote{#1}%
  \addtocounter{footnote}{-1}%
  \endgroup
}

\begin{document}

\title{Mapping Auto-context Decision Forests to Deep ConvNets for Semantic Segmentation}

\titlerunning{Mapping Auto-context to Deep ConvNets}        

\author{David L. Richmond\textsuperscript{*} \and
	    Dagmar Kainmueller\textsuperscript{*} \and
		Michael Y. Yang \and
		Eugene W. Myers \and
		Carsten Rother
}

\authorrunning{Richmond and Kainmueller et al.} 

\institute{
		David Richmond \at
			  Image and Data Analysis Core, Harvard Medical School, Boston, MA, USA 
			             \and
           Dagmar Kainmueller \at
              BIH/MDC, Berlin, Germany \\
			  dagmar.kainmueller@mdc-berlin.de
			             \and
			             Michael Yang \at
			                University of Twente, The Netherlands 
			                			  \and
			  Eugene Myers \at
			  MPI-CBG, Dresden, Germany 
			  			             \and
			             Carsten Rother \at
			                HCI/IWR, Heidelberg University, Germany 
			                }

\date{}

\maketitle

\begin{abstract}
We consider the task of pixel-wise semantic segmentation given a small set of labeled training images.
Among two of the most popular techniques to address this task are Decision Forests (DF) and Neural Networks (NN).
In this work, we explore the relationship between two special forms of these techniques: stacked DFs (namely Auto-context) and deep Convolutional Neural Networks (ConvNet).
Our main contribution is to show that Auto-context can be mapped to a deep ConvNet with novel architecture, and thereby trained end-to-end.
This mapping can be used as an initialization of a deep ConvNet, enabling training even in the face of very limited amounts of training data.
We also demonstrate an approximate mapping back from the refined ConvNet to a second stacked DF, with improved performance over the original.
We experimentally verify that these mappings outperform stacked DFs for two different applications in computer vision and biology:  Kinect-based body part labeling from depth images, and somite segmentation in microscopy images of developing zebrafish.
Finally, we revisit the core mapping from a Decision Tree (DT) to a NN, and show that it is also possible to map a fuzzy DT, with sigmoidal split decisions, to a NN.
This addresses multiple limitations of the previous mapping, and yields new insights into the popular Rectified Linear Unit (ReLU), and more recently proposed concatenated ReLU (CReLU), activation functions.
\blfootnote{$^*$ Shared first authors}
\keywords{Pre-training \and Sparse Convolutional Kernel \and Fuzzy Decision Tree \and Concatenated Rectified Linear Unit}
\end{abstract}

\section{Introduction}
\label{intro}

Deep learning has transformed the field of computer vision, and now rivals human-level performance in tasks such as image classification~\cite{krizhevsky_cnn_2012, He:2015dj, Russakovsky:2015hb}, facial recognition~\cite{Taigman:2014gy}, and object detection~\cite{GirshickDDM14, Girshick:2015vr, Ren:2015ug}. These advances have been fuelled by large labeled data sets, such as ImageNet \cite{Russakovsky:2015hb}, that can be used to train high capacity, deep ConvNets. Once trained, these models serve as generic feature extractors that can be applied to a wide range of problems using simple refinement techniques~\cite{GirshickDDM14}. An example of this is semantic segmentation by a Fully Convolutional Network that was pre-trained for image classification on ImageNet~\cite{long_shelhamer_fcn_2015}.

Despite the overwhelming success of the prescribed approach, there are still many specialized tasks that cannot be easily addressed by refining pre-trained networks, and for which there does not exist a sufficiently large data set to train a high capacity ConvNet from scratch. An important example of this is in biomedical imaging, where no AlexNet exists, and there is often a dearth of publicly available data.

\begin{figure*}
\begin{center}
   \includegraphics[width=1.0\textwidth]
                   {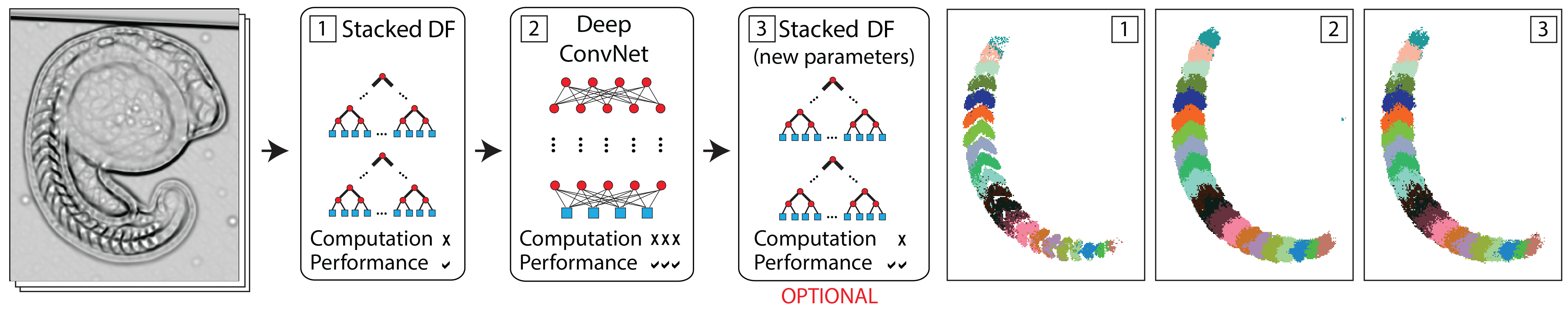}
\end{center}
   \caption{\textbf{Overview.} Our method (left) and corresponding results (right) for semantic segmentation of somites in microscopy images of developing zebrafish. (1) A stacked DF is trained to predict dense semantic labels from an input feature stack. (2) The stacked DF is then mapped to a deep ConvNet and further trained by back-propagation to improve performance. (3) Optionally, the ConvNet is mapped back to a stacked DF with updated parameters. The new stacked DF performs worse than the ConvNet but requires much less computation, and is better than the original DF}
\label{fig:pipeline}
\end{figure*}

A common strategy when training data is limited, is to use ensemble approaches, such as Decision Forest classifiers (DF).  The use of stacked classifiers, such as Auto-context \cite{Tu2010}, creates ``deep'' classifiers that have been shown to learn contextual information and thereby improve performance on many tasks such as object-class segmentation \cite{ShottonJC08}, facade segmentation \cite{jampani}, and brain segmentation \cite{Tu2010}. However, this strategy has the limitation that the stack of classifiers is trained greedily, in contrast to the end-to-end training of deep ConvNets, thereby limiting its performance. Thus, there is a need for methods that train stacked classifiers end-to-end. 
Our work addresses this issue by exploiting the connection between DTs and NNs~\cite{Sethi1990}.
Figure~\ref{fig:pipeline} depicts our proposed pipeline.

\textbf{Contributions:}
 
1.         We show that a stacked DF with contextual features is a special case of a deep ConvNet with sparse convolutional kernels. 

2.		   We describe a mapping from a stacked DF to a sparse, deep ConvNet, and utilize this mapping to initialize the ConvNet from a pre-trained stacked DF. This leads to superior results on semantic segmentation with limited training samples, compared to alternative strategies.
 
3.         We describe an approximate mapping from our sparse, deep ConvNet back to a stacked DF with updated parameters, for more computationally efficient evaluation, \textit{e.g.}, for low power devices. We show that this improves performance as compared to the original stacked DF.
 
4.         Due to our special ConvNet architecture we are able to gain new insights into the activation pattern of internal layers, with respect to semantic labels. In particular, we observe that the common smoothing strategy in stacked DFs is naturally learned by our ConvNet.

5.		   We revisit the core mapping from a DT to a NN, and show that it is possible to map a fuzzy DT, with sigmoidal split decisions, to a NN.  This mapping addresses some limitations of the previously described mapping, and gives a new interpretation of the recently proposed concatenated ReLU activation.

\section{Related Work}

Our work relates to (i) global optimization of DF classifiers, (ii) feature learning in stacked DF models, and (iii) applying ConvNets to the task of semantic segmentation.

\textbf{Global Optimization of DFs.} The limitations of traditional greedy DF construction~\cite{breiman_2001} have been addressed by numerous works.
In~\cite{SuarezL99}, the authors learn DTs by the standard method (see \cite{Criminisi:2013us} for a detailed description of this method), followed by a process called ``fuzzification'', replacing all threshold split decisions with smooth sigmoid functions, and derive a tree-based back-propagation algorithm for jointly refining split parameters.
In~\cite{Norouzi}, they derive a convex-concave upper bound as a proxy for a global loss function, and use this to jointly train all split functions and leaf parameters by stochastic gradient descent.  However, in contrast to~\cite{SuarezL99}, they impose constraints to preserve hard split decisions, such that each sample traverses a single path to a single leaf in each tree.
In~\cite{shaoqing15grrf}, they focus on combining the predictions from each DT so that the complementary information between multiple trees is optimally exploited with respect to a final loss function.
After training a standard DF, they retrain the distributions stored in the leaves, and prune the DTs to accomplish compression and avoid overfitting.
A related approach is to train an DF, and then map to a shallow NN with two hidden layers and refine the parameters by back-propagation. 
This was originally demonstrated for classification \cite{Sethi1990,Welbl14}, and more recently for regression \cite{welbl:regression}.
As opposed to \cite{SuarezL99} and \cite{shaoqing15grrf}, this enables end-to-end training of all parameters with respect to a final loss function.
Our work builds upon \cite{Sethi1990,Welbl14}: We extend their approach to a deep ConvNet, inspired by the Auto-context algorithm \cite{Tu2010}, and apply it to semantic segmentation.

\textbf{Feature Learning in a DF Framework.}
Auto-context introduces new contextual features during the learning process, and thus is a form of feature learning.
Numerous works have generalized this approach.
In Entangled Random Forests (ERFs)~\cite{ERFs}, spatial dependencies are learned using ``entanglement features'' in each DT, without the need for stacking.
Geodesic Forests~\cite{KontschiederKSC13} apply image-aware geodesic smoothing to the class distributions, to generate features for deeper nodes in the DT.
However, 
these approaches are still limited by greedy parameter optimization.

In a more traditional approach to feature learning, Neural Decision Forests~\cite{BuloK14} mix DFs and NNs by using multi-layer perceptrons (MLP) as soft split functions, to jointly tackle the problem of data representation and discriminative learning.
This approach can obtain superior results with smaller trees, at the cost of more complicated split functions; however, the MLPs in each split node are trained independently of each other.
The authors in~\cite{DeepNDFs} address this limitation by training the entire system end-to-end, and this is the most closely related to our work; however, they adopt a mixed framework, with both DFs and ConvNets trained in an alternating fashion, and apply their model to the task of image classification.
By contrast, our work maps a stacked DF model to the ConvNet framework, which enables optimization with the standard back-propagation algorithm, and apply this model to the task of semantic segmentation.

\textbf{ConvNets for Semantic Segmentation.} Conv\-Nets can be applied to semantic segmentation either in a tile-based manner \cite{Ciresan}, or using ``whole-image-at-a-time'' processing in a Fully Convolutional Network (FCN) \cite{long_shelhamer_fcn_2015}.
A challenge of these approaches is that the built-in spatial invariance of ConvNets trained for image classification leads to coarse-graining effects on the output.
A variant of FCN called U-Net was proposed in~\cite{unet}, and uses skip layers to combat coarse-graining during up-sampling.
More recently, the authors in~\cite{Badrinarayanan:2015ub} propose non-linear upsampling, by making use of the pooling indices stored in the down-sampling layers of the ConvNet.
In \cite{Zheng, Arnab:2016uk}, they address coarse-graining by expressing mean-field inference in a dense CRF as a Recurrent Neural Network (RNN), and concatenating this RNN behind a FCN, for end-to-end training of all parameters. Notably, they demonstrate a significant boost in performance on the Pascal VOC 2012 segmentation benchmark; however, this model is trained on large scale data and has not been applied to scenarios, such as biomedical images, with limited training examples.

Unsupervised pre-training has been used successfully to leverage smaller labeled training sets \cite{Sermanet, Ranzato, Hinton:DeepBeliefNets}; however, fully supervised training on large data sets still gives higher performance.
A common practice is to train a ConvNet on a large training set, and then fine tune the parameters on the target data \cite{GirshickDDM14}; however, this requires a closely related task with a large labeled data set, such as ImageNet.
It has been shown that for domain specific tasks, the benefit of starting from a generic ImageNet trained model, relative to \textit{e.g.}, self-supervised pre-training, can be minimal~\cite{Sudowe:tc}.
Another strategy to address the dependency on training data, is to expand a small labeled training set through data augmentation \cite{unet}.

We propose a novel strategy for semantic segmentation with limited training data.
Similar to \cite{Fahlman, Lengelle}, we employ supervised pre-training, but in a complementary model, namely the Auto-context model \cite{Tu2010}.
Our approach avoids coarse-graining by generating a ConvNet architecture with no striding or pooling layers. 
Our method achieves a large receptive field with few parameters by using sparse convolutional kernels, similar to \cite{Chen, Yu:2015uc}; however, we learn the optimal position of the non-zero kernel element(s) during construction of the DF stack.
There has been recent interest in the use of sparse convolutional kernels for reducing computation in ConvNets \cite{Zisserman:lowRank, Criminsi:lowRank, Liu:sparseCNN}.  
Indeed ConvNets are known to be highly redundant and the number of parameters can in some cases be reduced by up to 90\% with only a 1\% loss in accuracy \cite{Liu:sparseCNN}.

\section{Method}
\label{sec:method}

Our contributions build upon the mapping of a DF to a NN with two hidden layers as proposed in~\cite{Sethi1990,Welbl14}.
In Section~\ref{subsec:welbl} we briefly review this mapping, adopting the notation of \cite{Welbl14} for consistency.
In Section~\ref{subsec:forward-map}, we describe our main contribution, namely how to map a stack of DFs onto a deep ConvNet.
In Section~\ref{subsec:back-map}, we describe our second contribution, an algorithm for mapping our deep ConvNet back onto the original DF stack, with updated parameters.

\subsection{Mapping a DF to a NN with Two Hidden Layers}
\label{subsec:welbl}

\begin{figure}
\begin{center}
\begin{tabular}{cc}

   \includegraphics[height=0.25\textwidth]
                   {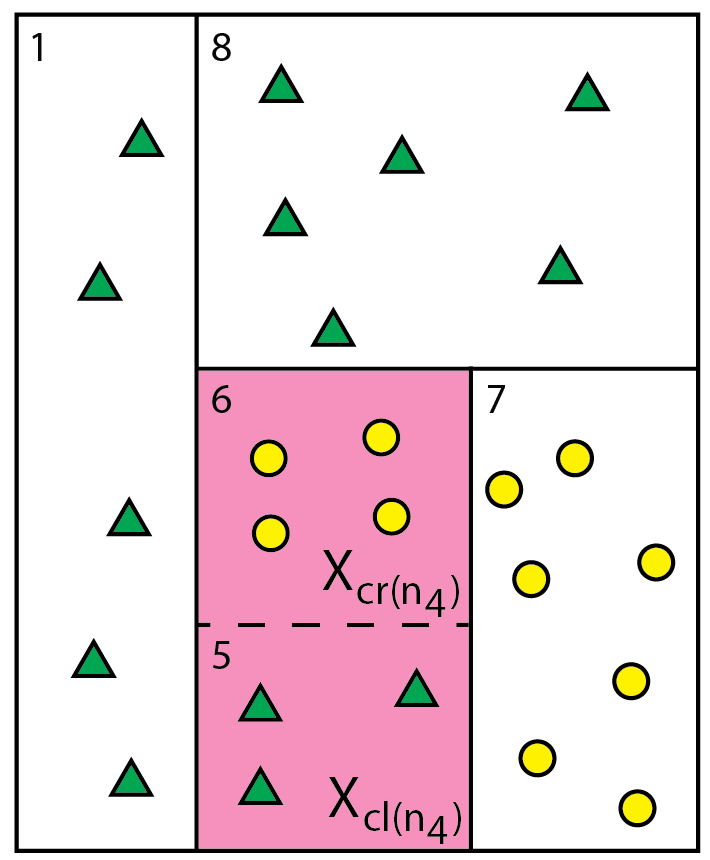} &
   \includegraphics[height=0.25\textwidth]
                   {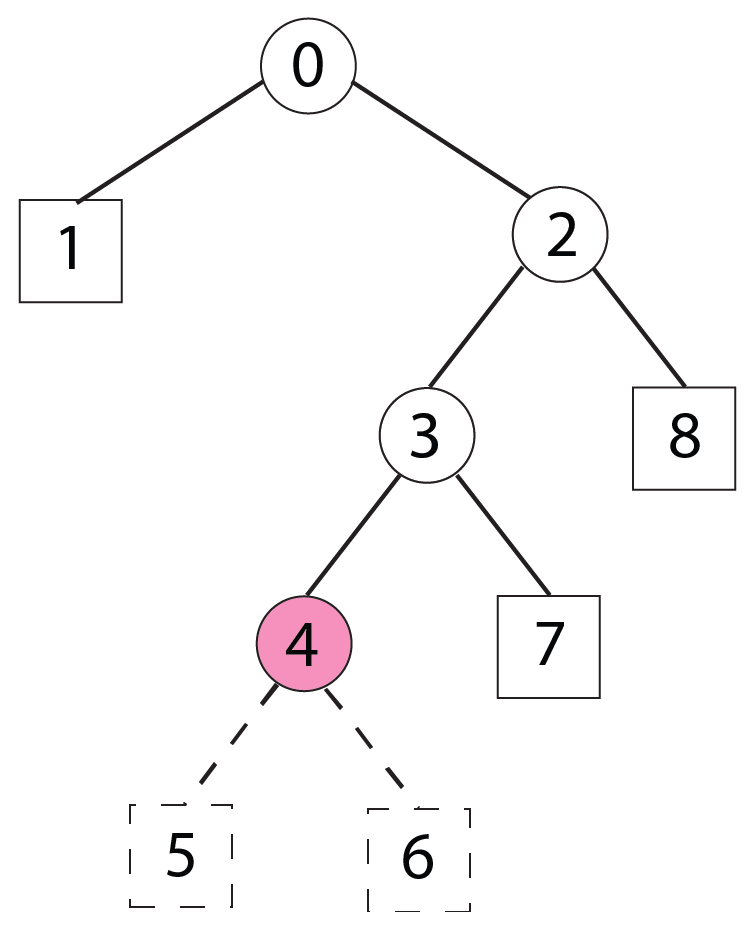} \\ \\
				   (a) & (b)
\end{tabular}
\end{center}
   \caption{\textbf{Partitioning of input feature space by a Decision Tree.} (a) Feature space, $X$, with samples, $\mathbf{x}_i$, from 2 classes, and (b) the corresponding DT.  Regions of feature space are numbered consistently with leaf node labeling in the tree.  As an example of feature space partitioning, split node $n_4$ processes the region $X_{n_4}$ shown in red, and splits it into sub-regions $X_{cl(n_4)}$ and $X_{cr(n_4)}$.}
\label{fig:RFpartitioning}
\end{figure}

A DT consists of a set of split nodes, $n \in \mathcal{N}^{Split}$, and leaf nodes, $l \in \mathcal{N}^{Leaf}$.
Each split node $n$ processes the subset $X_n$ of the feature space $X$ that reaches it. 
Usually, $X=\mathbb{R}^F$, where $F$ is the number of features. 
Let $cl(n)$ and $cr(n)$ denote the left and right \emph{child node} of a split node $n$.
A split node $n$ partitions the set $X_n$ into two sets $X_{cl(n)}$ and $X_{cr(n)}$ by means of a \emph{split decision}.
For DTs using axis-aligned split decisions, the split is performed on the basis of a single feature whose index we denote by $f(n)$, and a respective threshold denoted as $\theta_n$.  Thus, 
$\forall \mathbf{x} \in X_n: \mathbf{x} \in X_{cl(n)} \Longleftrightarrow x_{f(n)} < \theta_n $.  
See Figure~\ref{fig:RFpartitioning} for an illustration of this principle.

For each leaf node $l$, there exists a unique path from root node $n_0$ to leaf $l$, $P(l)=\{n_i\}_{i=0}^d$, with $n_0...n_d \in \mathcal{N}^{Split}$ and $X_l \subseteq X_{n_d} \subseteq ... \subseteq X_{n_0}$.
Thus, leaf membership can be expressed as follows:
\begin{equation}
\mathbf{x} \in X_l \Longleftrightarrow \forall n \in P(l) : \begin{cases}
x_{f(n)} < \theta_n & \text{if $X_l \subseteq X_{cl(n)}$}.\\
x_{f(n)} \geq \theta_n & \text{if $X_l \subseteq X_{cr(n)}$}.\\
\end{cases}
\label{eq:rf:leaf-membership}
\end{equation}
Each leaf node $l$ stores votes for the semantic class labels, $\mathbf{y}^l = (y^l_1...y^l_{C})$, where $C$ is the number of classes.
For a feature vector $\mathbf{x}$, we denote the unique leaf of the tree that has $\mathbf{x}\in X_l$ as $\leaf(\mathbf{x})$.
The prediction of a DT for feature vector $\mathbf{x}$ to be of class $c$ is given by:
\begin{equation}
p(c | \mathbf{x}) = \frac{y^{\leaf(\mathbf{x})}_c}{\sum_{c=1}^{C}y^{\leaf(\mathbf{x})}_c} 
\label{eq:dt:prediction}
\end{equation}

\begin{figure*}
\begin{center}
\begin{tabular}{ccc}

   \includegraphics[height=0.24\textwidth]
                   {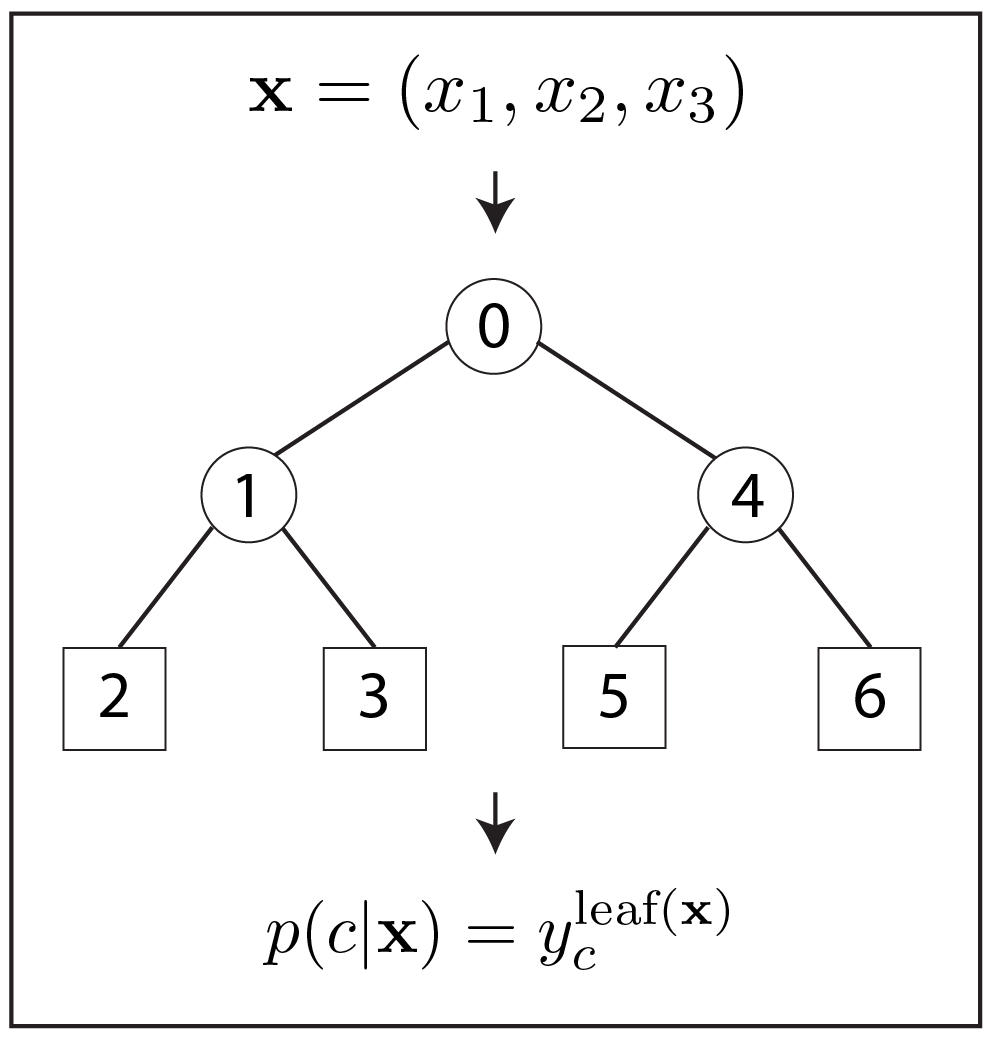} &
   \includegraphics[height=0.24\textwidth]
                   {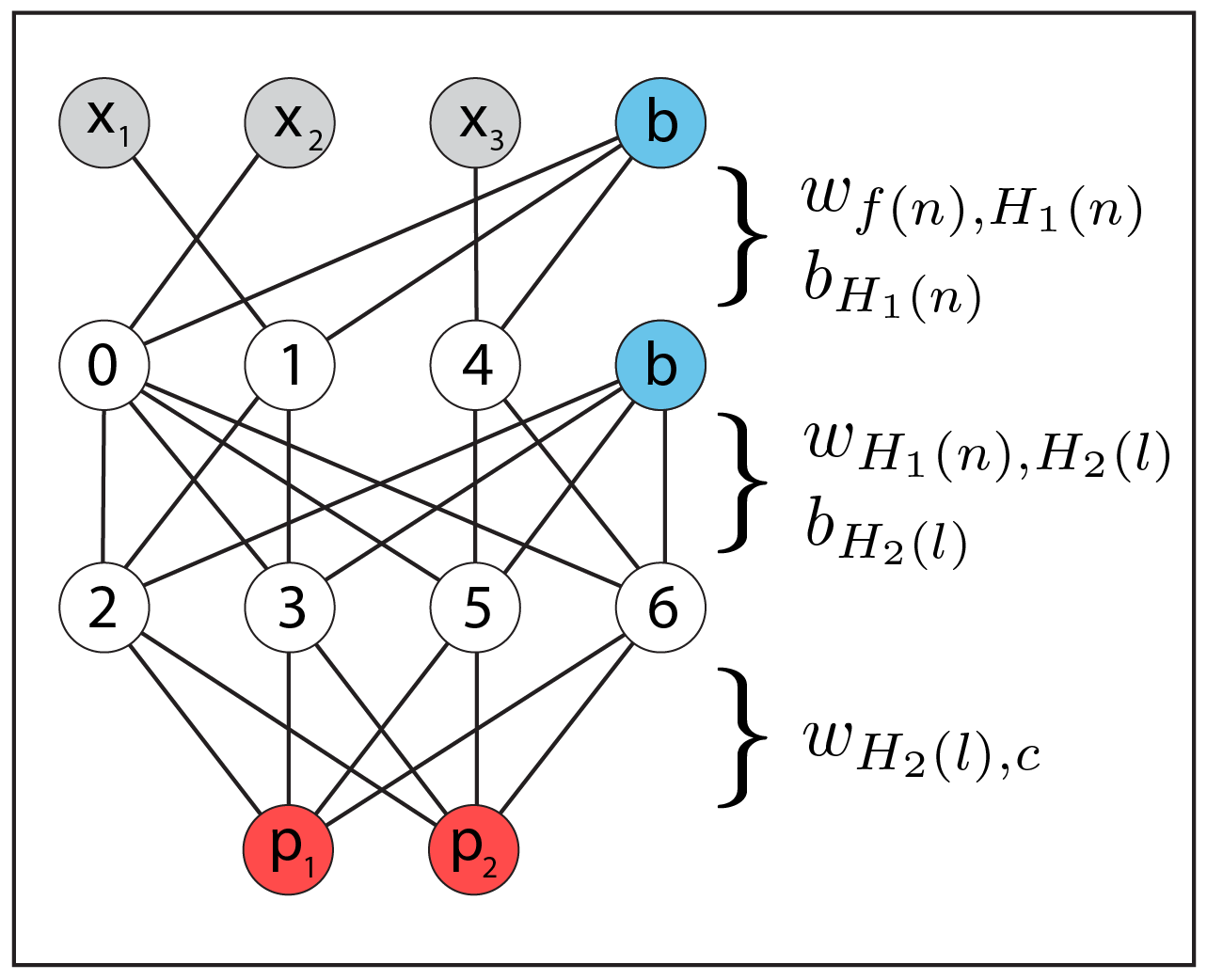} &
   \includegraphics[height=0.24\textwidth]
                   {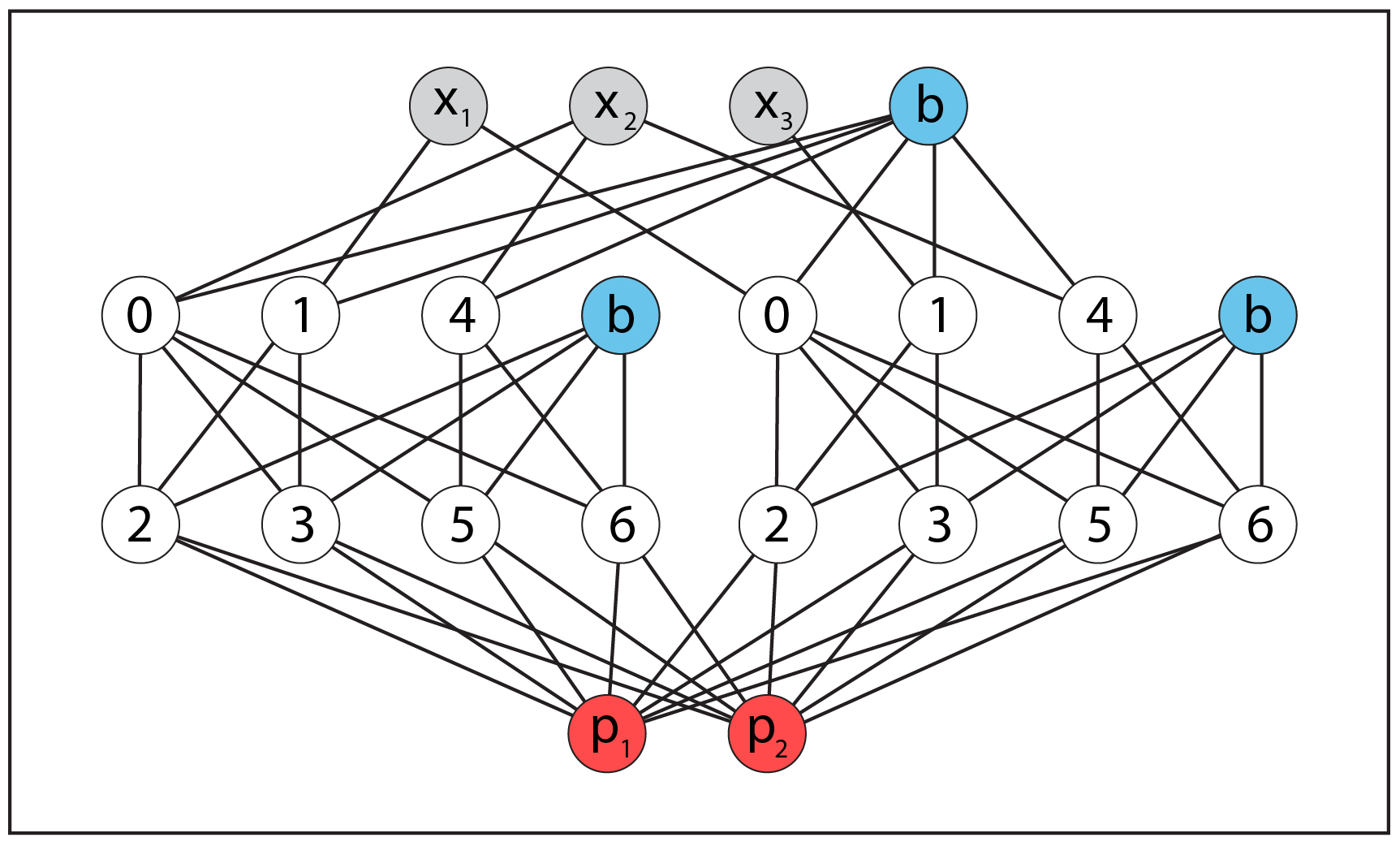} \\
				   (a) & (b) & (c)
\end{tabular}
\end{center}
   \caption{\textbf{Mapping from a DF to a NN.} (a) A shallow DT with input feature $\mathbf{x}$ represented by feature vector $(x_1,x_2,x_3)$. Nodes are labeled to show mapping to NN. (b) Corresponding NN with two hidden layers. The first hidden layer is connected to the input layer through weights $w_{f(n),H_1(n)}$, where $f(n)$ is the feature evaluated by split node $n$. \textit{E.g.}, $f(n_0)=2$.  This layer outputs the split decision for each split node of the DT (numbered 0,1,4). The weights $w_{H_1(n),H_2(l)}$ between the two hidden layers encode the structure of the tree. In particular, the split nodes along the path to leaf $l$ are connected to $H_2(l)$. For example, leaf node 5 is connected to split nodes 0 and 4, but not split node 1. The second hidden layer encodes leaf membership for each leaf node (numbered 2,3,5,6). The final weights $w_{H_2(l),c}$ are fully connected and store the votes $y^l_c$ for each leaf $l$ and class $c$. Gray: Input feature nodes. Blue: Bias nodes. Red: Prediction nodes, $p(c | \mathbf{x})$. (c) NN corresponding to a DF with two DTs, each with the same architecture as in (a). Note that, while the two DTs have the same architecture, they use different input features at each split node, and do not share weights}
\label{fig:forward-map_TREE}
\end{figure*}

Using this notation, we now describe how to map a DT to a feed-forward NN, with two hidden layers.
Conceptually, the NN separates the task of evaluating the split nodes and evaluating leaf membership into the first and second hidden layers, respectively. 
See
Figure~\ref{fig:forward-map_TREE}
for a sketch of the following description.

\textbf{Hidden Layer 1. } 
The first hidden layer, $H_1$, is constructed with one neuron, $H_1(n)$, per split node in the corresponding DT. This neuron evaluates $x_{f(n)} \geq \theta_n$, and encodes the outcome in its activity, $a_{H_1(n)}$. $H_1$ is connected to the input layer with weights and biases $w_{{f(n)},H_1(n)} = \alpha_1$ and $b_{H_1(n)} = -\alpha_1 \cdot \theta_n$.
Recall, $f(n)$ specifies the feature evaluated by split node $n$, and $\alpha_1$ sets how rapidly the neuron activation changes as its input crosses its threshold.
All other weights in this layer are zero.
Note, modeling DTs with oblique split functions simply results in multiple non-zero weights per neuron in this layer.

As activation function, $a_{H_1(n)}=\tanh(z_{H_1(n)})$ is employed, where $z_{H_1(n)} = \sum_{i} w_{i,H_1(n)}x_{i} + b_{H_1(n)}$ is the input to neuron $H_1(n)$ before non-linearity is applied.
A large value is used for $\alpha_1$ to approximate thresholded split decisions.
During training, $\alpha_1$ can be reduced to avoid the problem of diminishing gradients in back-propagation; however, for now we assume $\alpha_1$ is a large positive constant.
Thus, the pattern of activations encodes leaf node membership as follows:
\vspace{-6pt}

{
\small
\begin{equation}
\mathbf{x} \in X_l \Longleftrightarrow \forall n \in P(l) : \begin{cases}
a_{H_1(n)} = -1 \ \ \textnormal{if} \ \ X_l \subseteq X_{cl(n)} \\
a_{H_1(n)} = +1 \ \ \textnormal{if} \ \ X_l \subseteq X_{cr(n)} \\
\end{cases}
\label{eq:nn:leaf-membership}
\end{equation}
}

\textbf{Hidden Layer 2. } The role of neurons in the second hidden layer, $H_2$, is to interpret the activation pattern a feature vector $\mathbf{x}$ triggers in $H_1$,  and thus identify the unique $\leaf(\mathbf{x})$. 
Therefore, for every leaf $l$ in the DT, one neuron is created, denoted as $H_2(l)$. Each such neuron is connected to all $H_1(n)$ with $n \in P(l)$, but no others. Weights are set as follows: 
\vspace{-6pt}

{
\small
\begin{equation}
w_{H_1(n),H_2(l)} = \begin{cases}
-\alpha_2 \ \ \textnormal{if} \ \ X_l \subseteq X_{cl(n)} \\
+\alpha_2 \ \ \textnormal{if} \ \ X_l \subseteq X_{cr(n)} \\
\end{cases}
\end{equation}
}

\vspace{-6pt}
\noindent The sign of these weights matches the pattern of incoming activations \emph{iff} $\textbf{x} \in X_l$, thus making the activation of $H_2(l)$ maximal.
To distinguish leaf membership, the biases in $H_2$ are set as:
\vspace{-6pt}

{
\small
\begin{equation}
b_{H_2(l)} = -\alpha_2 \cdot ( |P(l)| - 1) 
\end{equation}
}

\vspace{-6pt}
\noindent Thus the input to node $H_2(l)$ is equal to~$1$ if $\mathbf{x} \in X_l$, and less than or equal to $-1$ otherwise. 
Using sigmoid activation functions, and a large value for $\alpha_2$, the neurons approximately behave as binary switches that indicate leaf membership. \emph{I.e.}, $a_{H_2(\leaf(\mathbf{x}))} = 1$ and all other neurons are silent.

\textbf{Output Layer. } The output layer of the NN has $C$ neurons, one for every class label.  This layer is fully connected; however, there are no bias nodes introduced. The weights store scaled votes from the leaves of the corresponding DT: $w_{H_2(l),c} = \alpha_3 \cdot y^l_c$. A softmax activation function is applied, to ensure a probabilistic interpretation of the output after training:
\begin{equation}
p(c | \mathbf{x}) = \frac{exp(\alpha_3 \cdot y^{\leaf(\mathbf{x})}_c)}{\sum_{c=1}^{C} exp(\alpha_3 \cdot y^{\leaf(\mathbf{x})}_c)}
\end{equation} 
Note that the softmax activation slightly perturbs the output distribution of the original DF (cf. Equation~\ref{eq:dt:prediction}).
This can be tuned by the choice of $\alpha_3$, and in practice is a minor effect.
Importantly, the softmax activation preserves the MAP solution.

\textbf{From a Tree to a Forest. } 
Let the number of DTs in a forest be denoted as $T$. 
The prediction of a forest for feature vector $\mathbf{x}$ to be of class $c$ is the normalised sum over the votes stored in the single active leaf per tree $t$, denoted $\leaf_t(\mathbf{x})$:
\begin{equation}
p(c | \mathbf{x}) = \frac{\sum_{t=1}^{T} y^{\leaf_t(\mathbf{x})}_c}{\sum_{c=1}^{C}\sum_{t=1}^{T} y^{\leaf_t(\mathbf{x})}_c}
\label{eq:rf:prediction}
\end{equation} 
Extending the DT-to-NN mapping described above to DFs is trivial: (i) replicate the basic NN design $T$ number of times, once for each tree in the DF, and (ii) fully connect $H_2$ to the output layer
(see Figure \ref{fig:forward-map_TREE}(c)).
This accomplishes summing over the leaf distributions from the different trees, before the softmax activation is applied.

\begin{figure*}[htp!]
\begin{center}
\begin{tabular}{cc}
		
   \includegraphics[height=0.29\textwidth]
                   {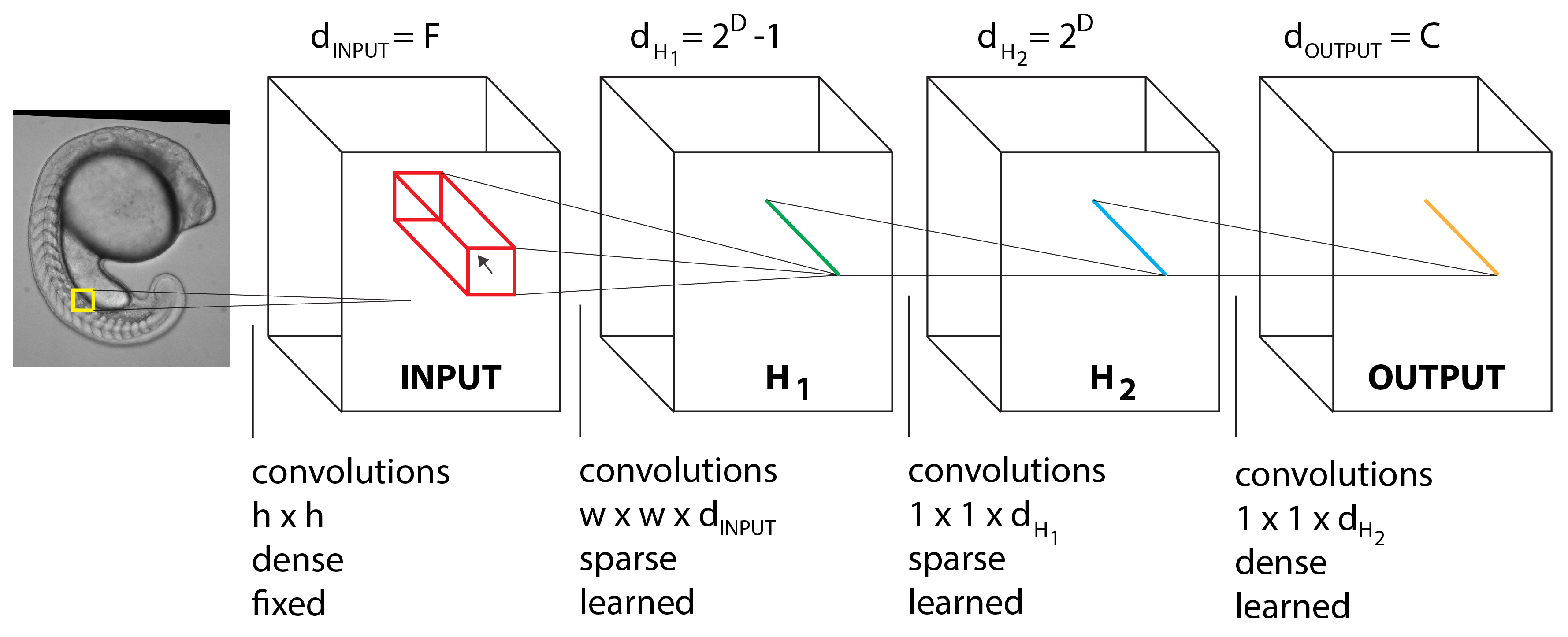} &
   \includegraphics[height=0.32\textwidth]
                   {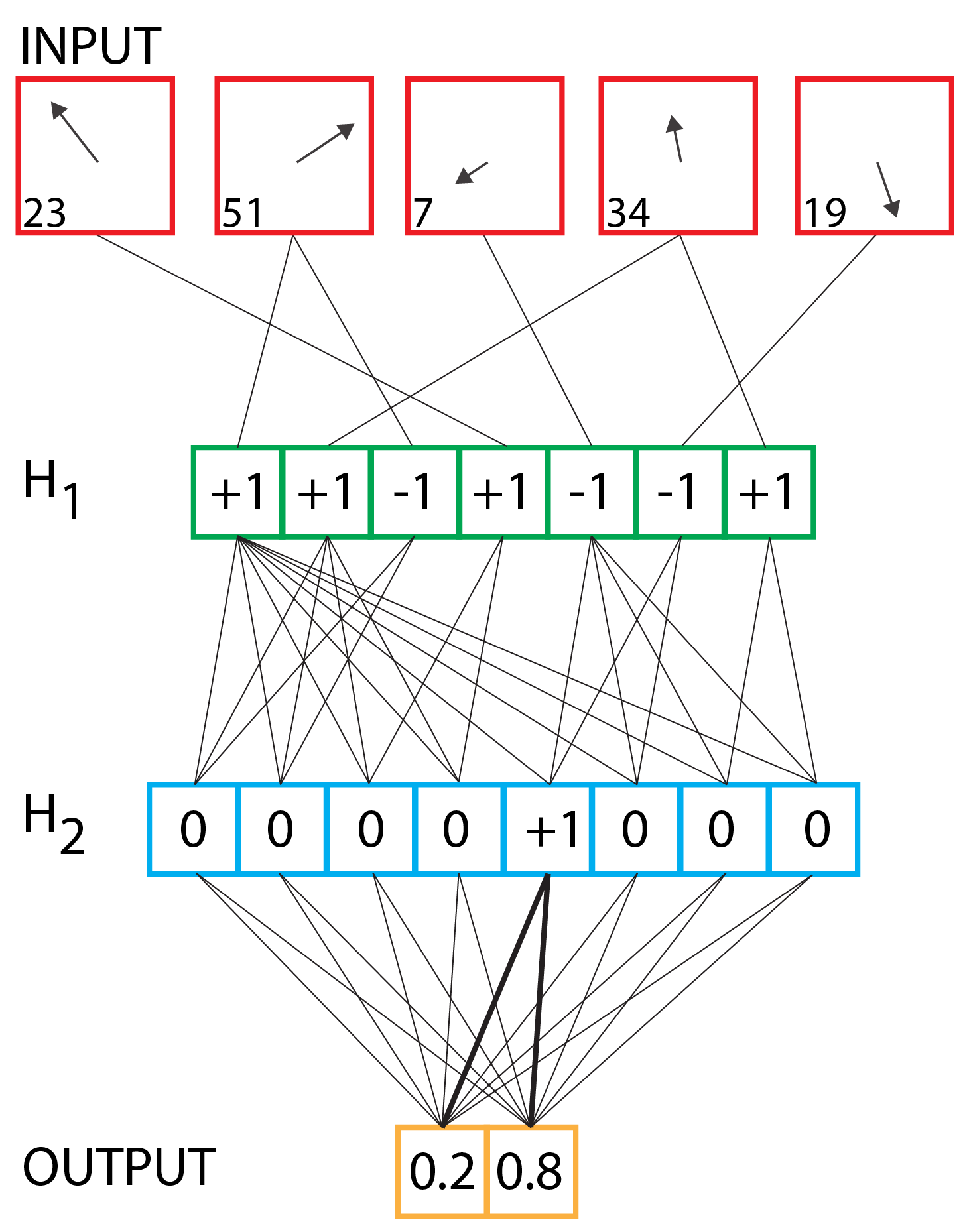} \\
				   (a) & (b)
\end{tabular}
\end{center}
   \caption{\textbf{ConvNet architecture of a DF. } (a) ConvNet architecture for dense semantic segmentation, corresponding to a DF with contextual features. The variables are, h: size of input convolution kernels, F: number of input convolution kernels, w: window size for offset features, d: number of feature maps in each layer, D: depth of corresponding DT, C: number of classes. (b) An example, where the DF is a single DT with depth D = 3, and 2 output classes. One pixel is classified in (b), corresponding to the region in (a) with similar color coding. The input layer (red) extracts features with a fixed offset (shown by arrows) and filter type (index into filter stack, shown at bottom left of each node). Activation values are shown for nodes in hidden layers $1$,$2$ and the output layer. In this example, the sample ends up in leaf $5$. Bias nodes are not shown for simplicity}
\label{fig:ConvNet}
\end{figure*}

We now discuss the relationship between DFs with contextual features and ConvNets.
In many applications such as body-pose estimation \cite{shaoqing15grrf}, medical image labeling \cite{ERFs}, and scene labeling \cite{Tu2010}, contextual information is included in the form of ``offset features'' that are selected from within a window defined by a maximum offset.
Such an DF can be viewed as a special case of a ConvNet, with sparse convolutional kernels and no max pooling layers (Figure~\ref{fig:ConvNet}).
The convolutions in hidden layer 1 have dimension $w$ x $w$ x $F$, where $w$ is the width of the offset window, and $F$ is the number of input features.
These kernels can be very sparse, \emph{e.g.,} it is common to have only a single non-zero element, or in the case of medical imaging, to use average intensity over a smaller offset window \cite{ERFs}.
The second layer convolutions have dimension $1$ x $1$ x $(2^D-1)$, where D is the depth of the DT, and are similarly very sparse, with only $D$ non-zero elements.

\subsection{Mapping a DF Stack to a Deep ConvNet}
\label{subsec:forward-map}

\begin{figure*}[htp!]
	\begin{center}
		\begin{tabular}{cc}
		
   		 	\includegraphics[height=0.45\textwidth]
                   {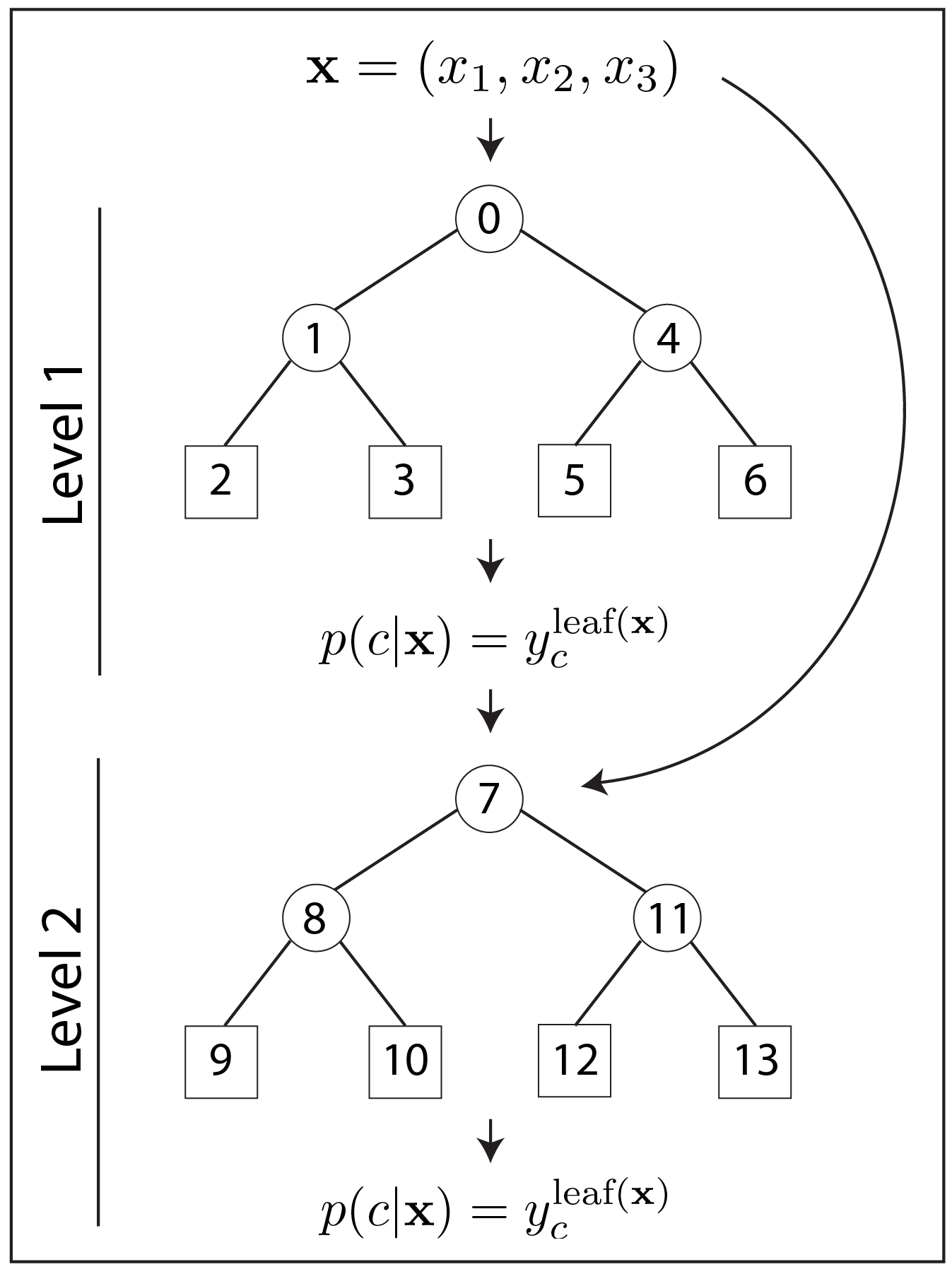} &
   			\includegraphics[height=0.45\textwidth]
                   {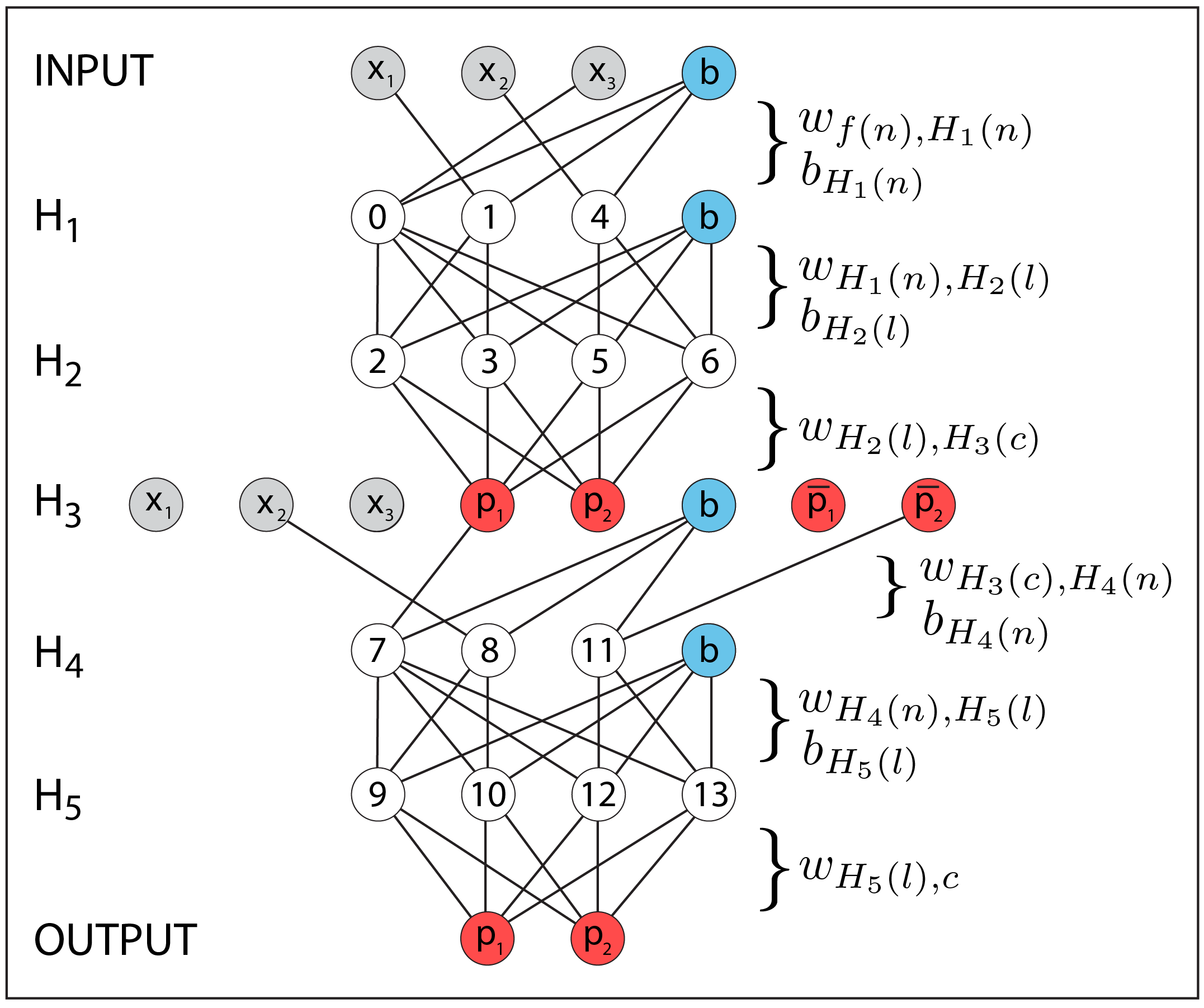} \\
				   (a) & (b)
		\end{tabular}
	\end{center}
    \caption{\textbf{Mapping from a stacked DF to a deep ConvNet.} (a) A stacked DF consisting of two shallow DTs. The second DF takes as input the original stack of convolutional filter responses, and the output of the previous DF across the entire window over which contextual features can be sampled. (b) Corresponding ConvNet with 5 hidden layers. Same color coding and node labeling as in Figure~\ref{fig:forward-map_TREE}.
In this example, the second DT learned to use filter response $x_2$, the DF output for class 1 at that same pixel (\emph{i.e.,} $p_1$), and the DF output for class 2 at some different offset pixel, denoted $\bar{p}_2$. Note that $\bar{p}_2$ is not a bias node; it is a contextual feature and its value depends on weights in previous layers}
	\label{fig:forward-map}
\end{figure*}

A stacked DF consists of multiple DF classifiers in a sequence, such that subsequent DF classifiers in the stack can use the predictions from the previous DF as input features (see Figure~\ref{fig:forward-map}a).
Training is done iteratively, from the first to the last DF in the stack (see \cite{Tu2010} for more details).
It was noted in the original Auto-context algorithm that it is important to allow the later DF classifiers to select features, not only from the output of the previous classifier, but also from the input feature stack.
Finally, to capture contextual information, these features are sampled with a learned offset.

We map this architecture onto a deep ConvNet as follows: each DF is individually mapped to a ConvNet, and then concatenated such that the layers corresponding to intermediate DF predictions become hidden layers, used as input to the next ConvNet in the sequence (Figure~\ref{fig:forward-map}b).
For a $K$-level DF stack, this generates a deep ConvNet with $3K-1$ hidden layers.
We also connect the input feature stack as bias nodes in hidden layers $H_{3k}$, $k = 1...K-1$, and introduce contextual (\emph{i.e.} offset) features with the sparse convolution kernels discussed above.
Due to the use of contextual features, individual pixels cannot be processed independently, but rather the complete image must be run through one level at a time (similar to the Auto-context algorithm), such that all features are available for the next level.
Finally, we remove the softmax activation from internal layers $H_{3k}$, $k = 1...K-1$, and normalize their output, to match the behaviour of the DF stack.

An interesting observation is that addition of trees to the DF and/or growing trees to a greater depth simply increases the width of the ConvNet, but is always restricted to a shallow architecture with only two hidden layers.
However, stacking DFs naturally increases the depth of the ConvNet architecture.

\subsection{Mapping the Deep ConvNet back to a DF Stack}
\label{subsec:back-map}

ConvNets are highly redundant~\cite{Liu:sparseCNN} and thus require a lot of additional computation, which may limit their applications \textit{e.g.} on low power devices~\cite{GolodetzSVVCAPK15, Massiceti:2016wg}.
We explore the possibility of mapping our deep ConvNet back to the more computationally efficient architecture of a stacked DF.
Given a ConvNet constructed from a $K$-level DF stack, the weights $w_{H_{3k-2}(n),H_{3k-1}(l)}$, with $k=1...K$, manifest the original tree structure.
Thus, keeping these weights and the corresponding biases, $b_{H_{3k-1}(l)}$, fixed during training allows the ConvNet to be mapped back onto the structure of the original DF stack.
For a single level stack, the mapping is as follows: (i) $\theta_n = -b_{H_1(n)}/w_{f(n),H_1(n)}$, where $\theta_n$ is the threshold for split node $n$, and (ii) $y^l_c = w_{H_2(l),c}$.
We refer to this as ``Map Back 1''.
When evaluating this DF, a softmax activation function needs to be applied to the output distribution to mimic inference in the ConvNet.
For deeper stacks, the output of \emph{each} DF must be post-processed with the corresponding activation function in the ConvNet, which in this paper is simple class normalization.

The above approach is appropriate if only a single leaf neuron fires in hidden layer $2$ for each sample. However, after training by back-propagation, this activation pattern will likely become distributed, and our mapping may not make optimal use of the learned parameter refinement.
Here, we propose a strategy to capture the distributed activation of the ConvNet.
For input $\mathbf{x}$ and class $c$, we would ideally like to store in $\leaf(\mathbf{x})$ of each DT, the following inner product:

\begin{equation}
y^{\leaf(\mathbf{x})}_c = z^{\mathbf{x}}_c := \nobreak{\sum_{l} a^\mathbf{x}_{H_2(l)} \cdot w_{H_2(l),c}}
\end{equation}

\begin{figure*}
\begin{center}
\begin{tabular}{ccc}

   \includegraphics[width=0.3\textwidth]
                   {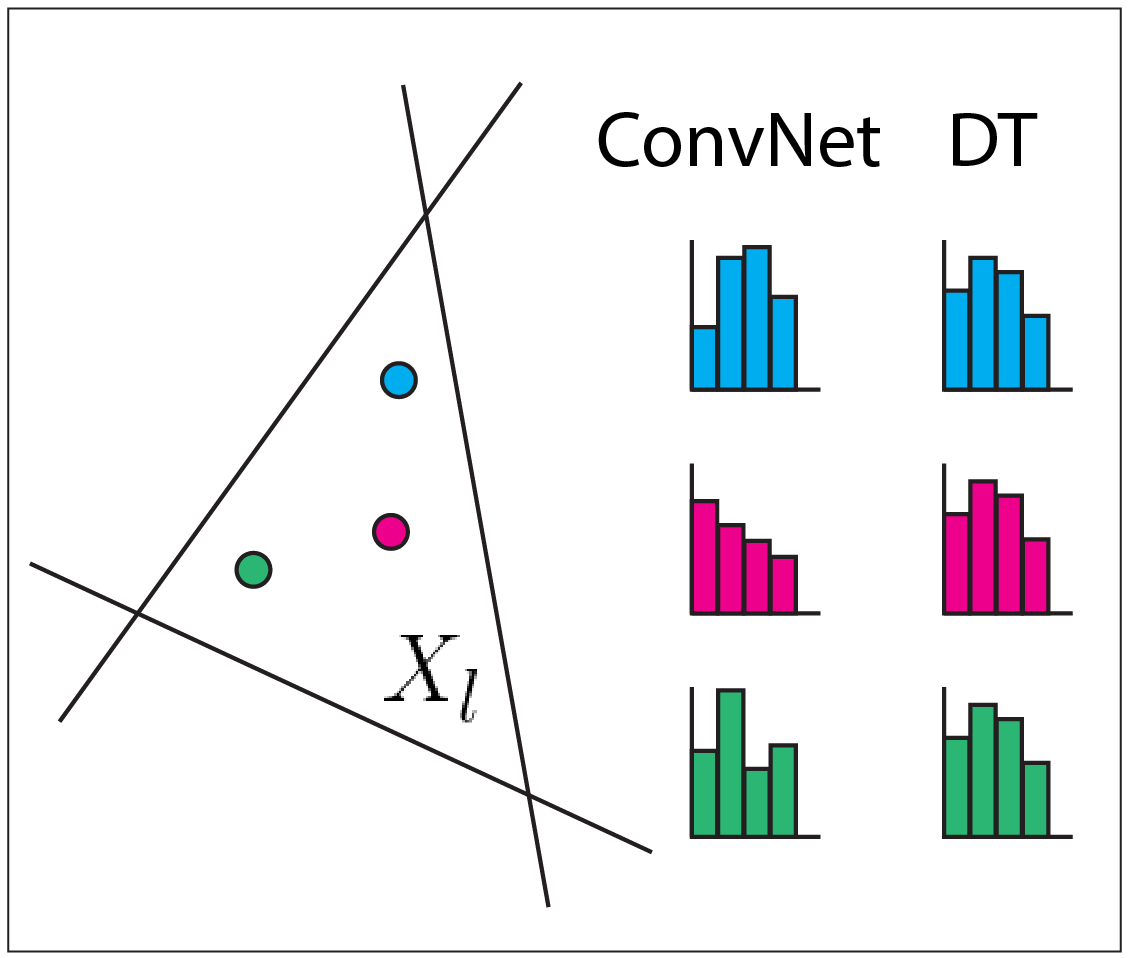} &
   \includegraphics[width=0.3\textwidth]
                   {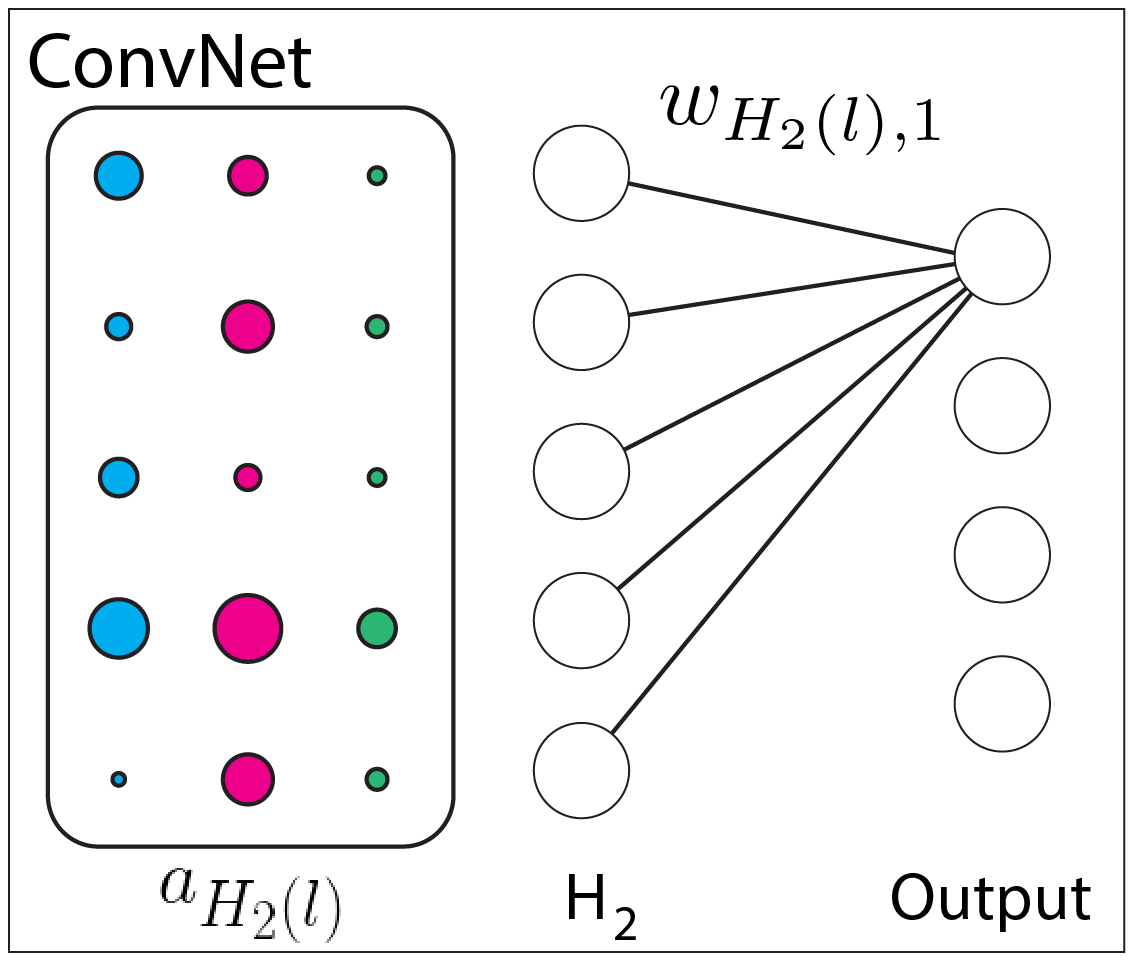} &
   \includegraphics[width=0.3\textwidth]
                   {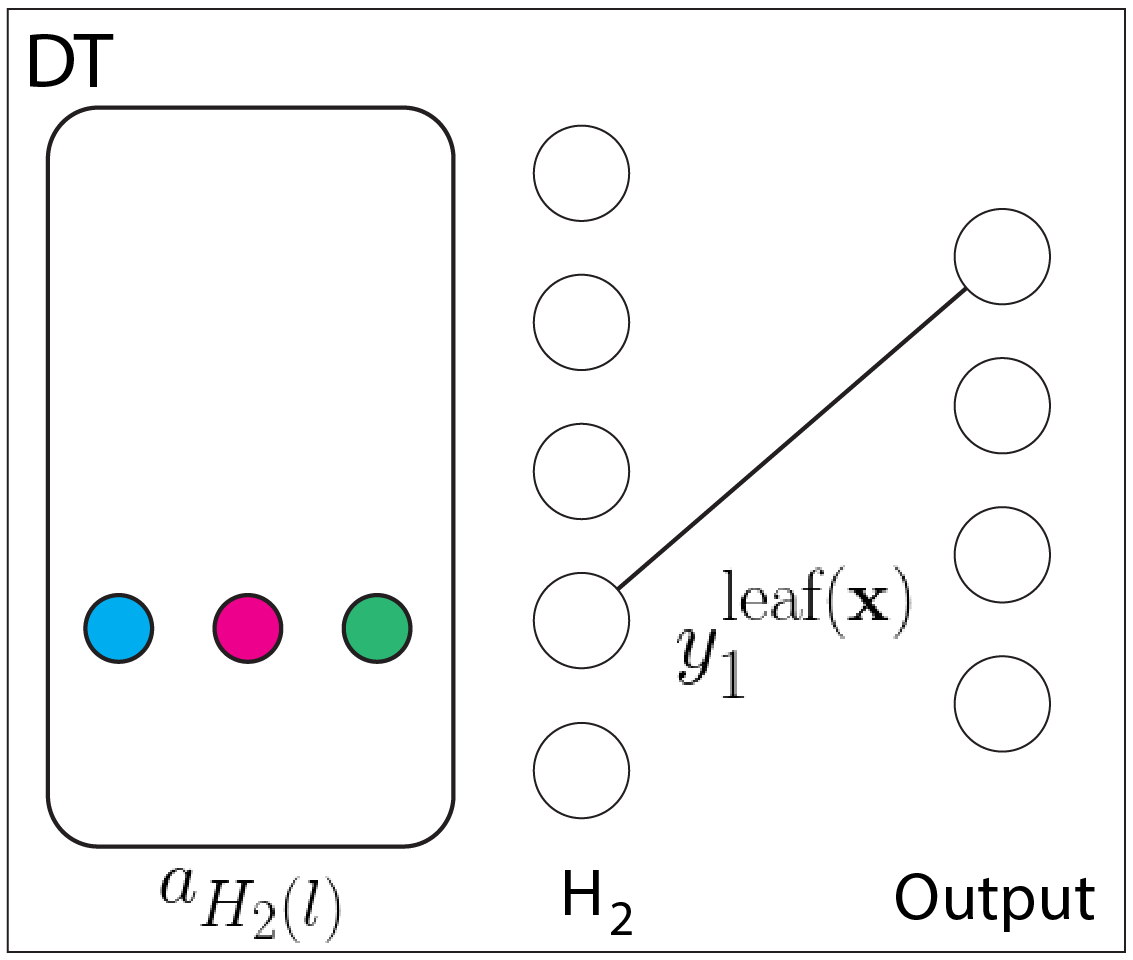} \\
				   (a) & (b) & (c) \\
\end{tabular}
\end{center}
   \caption{\textbf{Mapping ConvNet back to a DT. } (a)~Three samples (blue, magenta, green) falling into the leaf of a DT, corresponding to a subset of feature space, have the same posterior distributions; however, in a ConvNet their posteriors can be different. (b)~Corresponding activation pattern $a_{H_2(l)}$ for the three samples shown in (a) at hidden layer $2$ of the DF-initialized ConvNet. Radius of circles denotes the strength of the activation. The output layer receives the inner product of the activation pattern with weights $w_{H_2(l),c}$ (only weights to class $1$ shown for simplicity). (c)~Activation pattern in corresponding DT. Note, the inner product reduces to the value $y^l_1$ for class $1$. In Equation 2, we compute the optimal value of $y^l_c$, namely $\hat{y}^l_c$, to mimize the difference between the output of the DT and the ConvNet.}
\label{fig:map-back}
\end{figure*}

This mapping would elicit the identical output from the DF as from the ConvNet for input $\mathbf{x}$.
However, the activation pattern will vary for different training samples that end up in the same leaf, so this mapping cannot be satisfied simultaneously for the whole training set.
This results from the fact that DTs store distributions in their leaves that represent a piecewise-constant function on the input feature space, while the re-trained ConvNet allows for a more complex function
(see Figure~\ref{fig:map-back}).
As a compromise, we seek new vote distributions $\hat{y}^l_c$, for each $c,l,$ to minimise the following error, averaged over the finite set of training samples, $X^{train}$.

\begin{equation}
\hat{y}^l_c = \argmin_{y^l_c} \sum\limits_{\mathbf{x} \in X^{train}: \ \leaf(\mathbf{x})=l} \left( z^{\mathbf{x}}_c - y^l_c \right)^2
\label{eq:map-back:error}
\end{equation}
Equation~\ref{eq:map-back:error} can be solved analytically, yielding the following result:
\begin{equation}
\hat{y}^l_c = \frac{1}{N_l} \sum\limits_{\mathbf{x} \in X^{train} : \ \leaf(\mathbf{x})=l} z^{\mathbf{x}}_c 
\label{eq:map-back:weights}
\end{equation}

$N_l = |\{{\mathbf{x} \in X^{train}: \ \leaf(\mathbf{x})=l}\}|$, is the number of samples that end up in leaf $l$ of the DT.
Equation~\ref{eq:map-back:weights} is a simple average of $z^{\mathbf{x}}_c$ over all samples that end up in the same leaf.
We refer to this as ``Map Back 2''.
To implement this algorithm in a stack, we start by determining leaf membership for every sample and every tree in the first level of the DF stack. We then update the votes according to Equation~\ref{eq:map-back:weights}.  This is then repeated for all levels of the stack
(see Algorithm~\ref{alg:reweighting} for more details).
In the trivial case where, for every sample, a single neuron fires with unit activation in layers $H_{3k-1}$, $k=1...K$, this is equivalent to ``Map Back 1''.

\begin{algorithm}
\caption{\textbf{Mapping deep ConvNet back to K-level stacked DF.} The following algorithm was used to map the parameters from a trained ConvNet back to the original stacked DF architecture, and is referred to as Map Back 2.
We applied this algorithm to the zebrafish data set (Figure~\ref{fig:pipeline}: panel 3, and Figure~\ref{fig:comparison}(f)).}
\begin{algorithmic}
\State 1. Push all training data through ConvNet
\State 2. Store activations $a_{H_{3k-1}(l)}$, for $k = 1...K$
 \For{\texttt{ $i = 1 : K$}}
        \State Push all training data through stacked DF to level $i$
        \State Store $\leaf_t(\mathbf{x})$, for every tree $t$ and sample $\mathbf{x}$, at level $i$
        \State Update votes in $i^{th}$ DF to $\hat{y}^l_c$, according to Equation~\ref{eq:map-back:weights}
      \EndFor
\end{algorithmic}
\label{alg:reweighting}
\end{algorithm}

\section{Results}

The forward and backward mappings described above were implemented in Matlab, and tested on two different applications: Kinect-based body part labeling from depth images, and somite segmentation in microscopy images of developing zebrafish.

\subsection{Kinect Body Part Classification}

\begin{figure}
	\begin{center}
	  \begin{tabular}{ccccc}
		\includegraphics[height=0.24\textwidth]
            {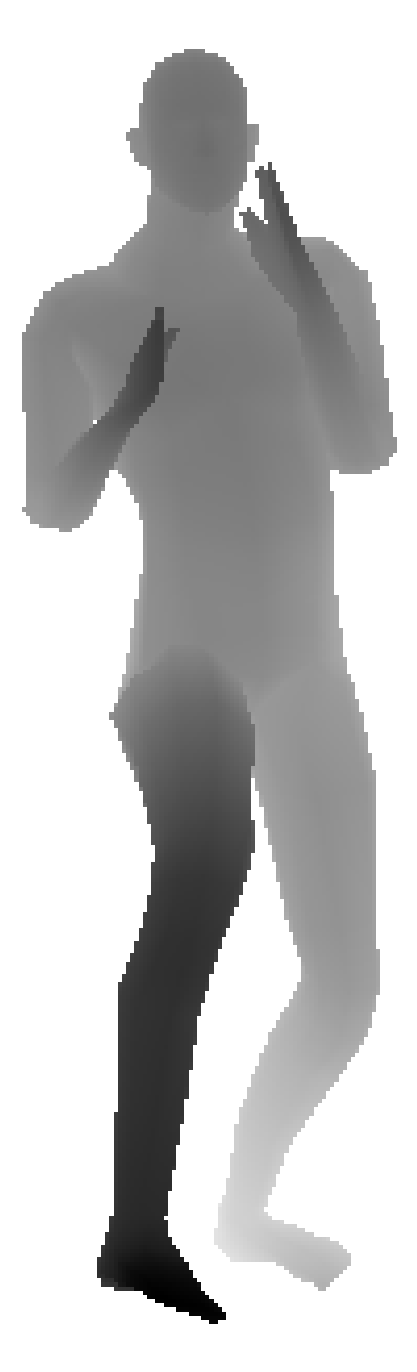} &
		\includegraphics[height=0.24\textwidth]
            {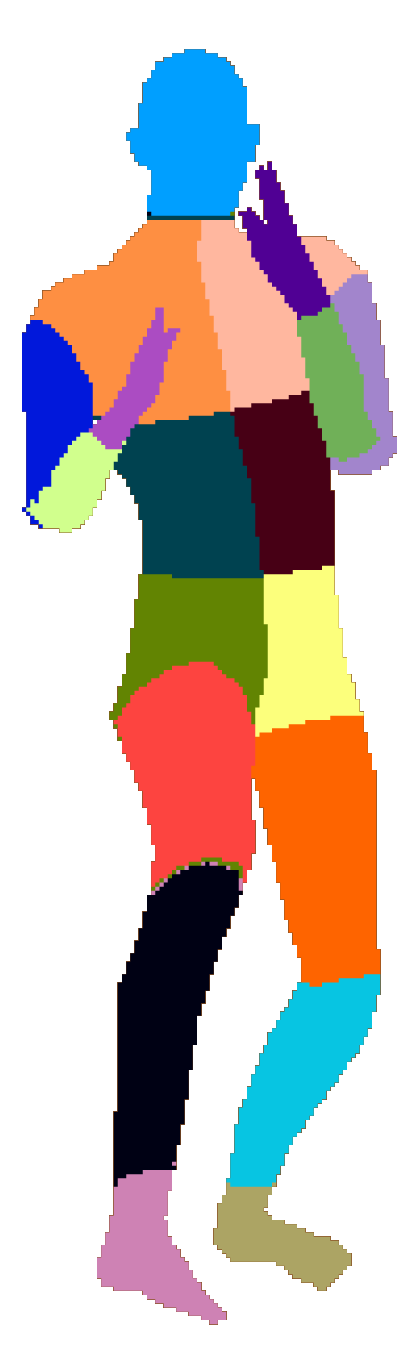} &
		\includegraphics[height=0.24\textwidth]
            {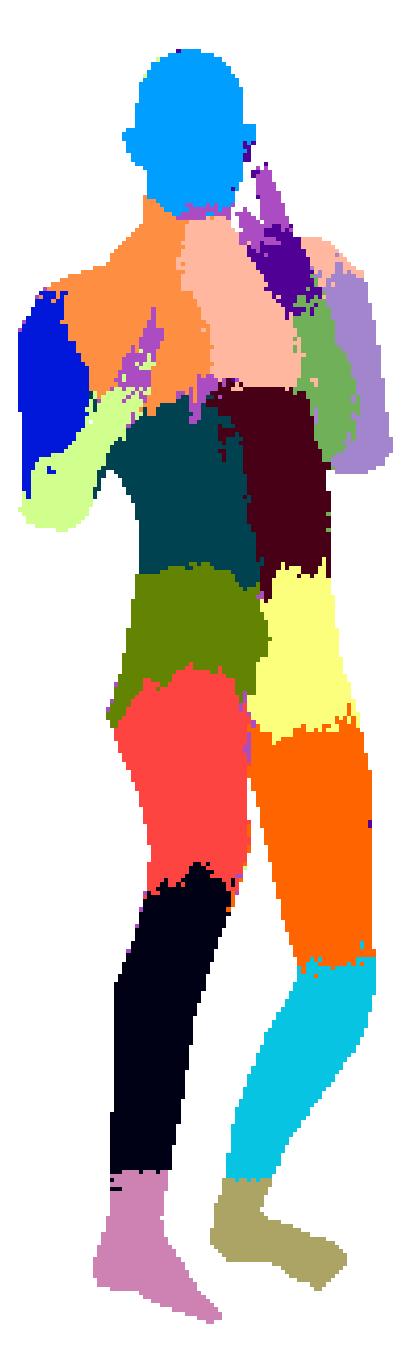} &
		\includegraphics[height=0.24\textwidth]
            {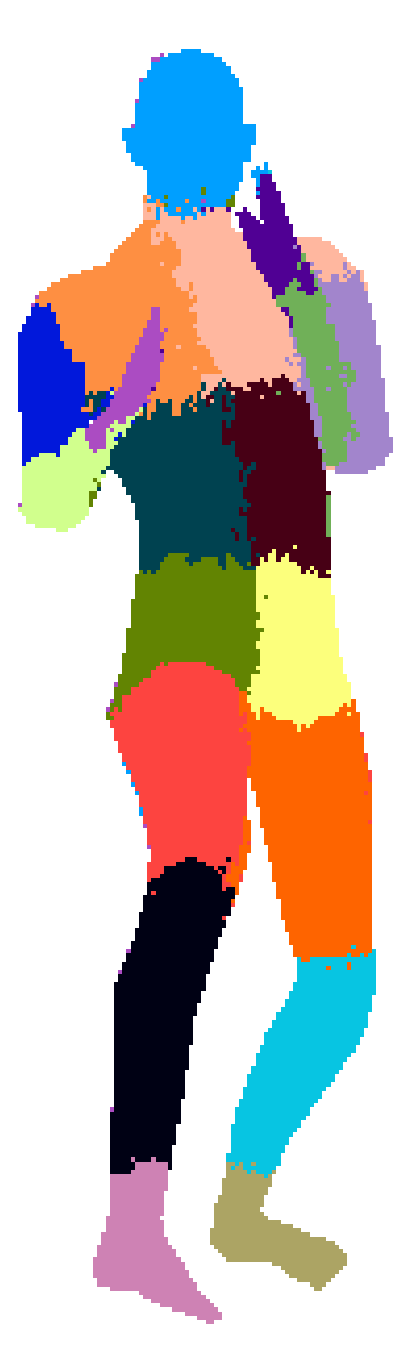} &
		\includegraphics[height=0.24\textwidth]
            {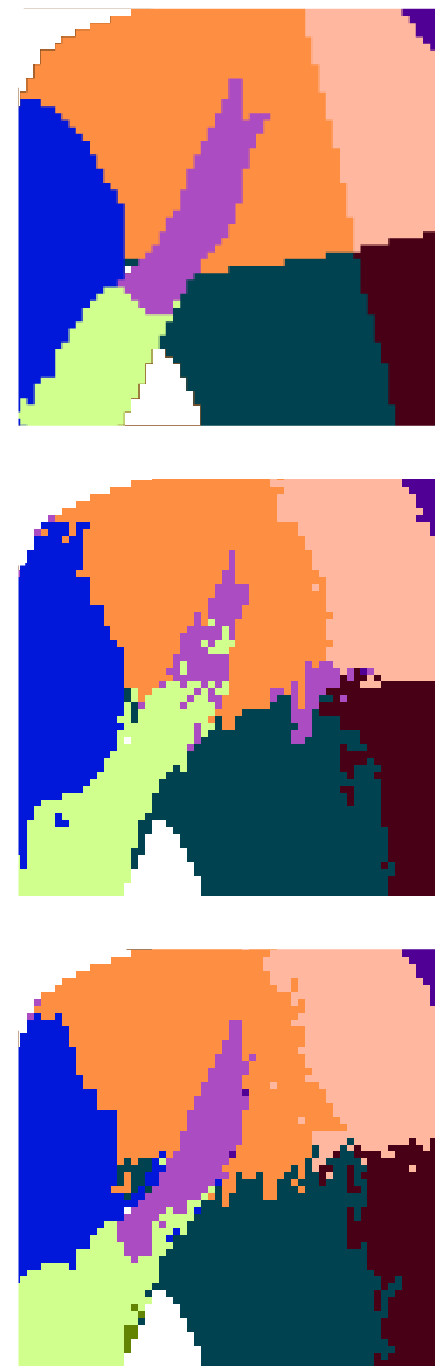} \\
			(a) & (b) & (c) & (d) & (e)
	  \end{tabular}
	\end{center}
	  \caption{\textbf{Example result of Kinect body part classification.} (a) Depth image. (b) Ground truth labels. (c) Result of stacked DF. (d) Result of DF-initialized ConvNet, after re-training. The accuracy for this test image increases from 0.88 to 0.94 on foreground classes. (e) Crop of hands for GT, DF and ConvNet, from top to bottom}
      \label{fig:kinect}
\end{figure}

\textbf{Experimental Setup. }
We applied our method to human body part classification from Kinect depth images, a domain where DFs have been highly successful \cite{shotton}.
We use the recently provided data set in \cite{denil}, since there is no publicly available data set from the original paper \cite{shotton}.
It contains 2000 training images, and 500 testing images, each 320x240 pixels, containing 19 foreground classes and 1 background class (Figure~\ref{fig:kinect}(a,b) for an example).
We evaluate the pixel accuracy, averaged over all foreground classes, as was done by \cite{denil}.
Note that background is trivially classified.

\textbf{Training of Stacked DF. }
We trained a two-level stacked DF, with the following parameters at every level: 10 trees, maximum depth 12, stop node splitting if less than 25 samples.
We selected 20 samples per class per image for training, and used the scale invariant offset features from \cite{shotton}, with standard deviation, $\sigma$ = 50 in each dimension.
Each split node selected the best from a random sample of 100 such features.

\textbf{Training of ConvNet. } 
We mapped the DF stack to a deep ConvNet with 5 hidden layers, as described in Section~\ref{sec:method}. For efficient training, the global parameters influencing the sharpness of the $\tanh$ activation functions were reduced such that the network could transmit a strong gradient via back-propagation. However, softening these parameters moves the deep ConvNet further from its initialization by the equivalent stacked DF. We evaluated a range of initialization parameters and found
$\alpha_1=\alpha_{4}=\alpha_{7}=100$, $\alpha_{2}=\alpha_{5}=\alpha_{8}=1$, $\alpha_{3}=\alpha_{6}=\alpha_{9}=0.1$ to be a good compromise, where $\alpha_{j}$ is the multiplicative factor applied to weights and biases in layer, $H_j$.

We trained the ConvNet with stochastic gradient descent (SGD) with momentum, and a cross-entropy loss.
We maintained the sparse connectivity from DF initialization, allowing only the weights on pre-existing edges to change, corresponding to the \textit{sparse} training scheme from~\cite{Welbl14}.

SGD training is applied by passing images through the network one at a time, and computing the gradient averaged over all pixels (\emph{i.e.}, batch size = 1 image).
Thus, we do ``whole-image-at-a-time'' training, as in \cite{long_shelhamer_fcn_2015}.
Since the network is designed for whole-image inputs, we first cropped the training images around the region of foreground pixels, and then down-sampled them by 25x.
Learning rate, $r$, was set such that for the $i^{th}$ iteration of SGD, $r(i) = a(1 + i/b)^{-1}$ with hyper-parameters $a = 0.01$ and $b = 400$ iterations.
Momentum, $\mu$, was set according to the following schedule: $\mu =\nobreak \mathtt{min} \{ \mu_{max}, 1 - 3 / (i + 5) \}$, where $\mu_{max} = 0.95$ \cite{sutskever}.
We trained for 8000 iterations, which takes approximately $10$ hours in our CPU-based Matlab implementation.

\textbf{Results. }
With our initial two-level stacked DF, we achieved a pixel accuracy of $0.82$, comparable to the original result of $0.79$ \cite{denil} (Figure~\ref{fig:kinect}(c), Table~\ref{tab:dice-score}(DF)).
After mapping to a deep ConvNet and re-training, we achieved an accuracy of $0.91$, corresponding to an $11\%$ relative improvement over the DF stack (Figure~\ref{fig:kinect}(d), Table~\ref{tab:dice-score}(ConvNet)).
This final result is comparable to the state-of-the-art result on this data set which aims to compress DFs by learning a better combination of their constituent trees \cite{shaoqing15grrf}.
They achieve a class-balanced pixel accuracy of $0.92$ over all classes, including the background class, for a model size of 6.8MB. Our model is smaller, at 3.3MB, due to our use of fewer and shallower trees.
Furthermore, they report that their optimization procedure takes multiple days, compared to our overnight refinement by back-propagation.
However, due to the different error metric, and their evaluation on a selected subset of pixels, the results are not directly comparable.

We also tried mapping the ConvNet back to the initial stacked DF architecture with updated parameters.
We first employed the trivial approach of mapping weights directly onto votes, similar to what was done in the DF to NN forward mapping; however, this reduced the Dice score to $0.74$ (Table~\ref{tab:dice-score}(MB1)), worse than the performance of the initial DF.
Next we applied Algorithm~\ref{alg:reweighting}, which yielded a final Dice score of $0.85$ (Table~\ref{tab:dice-score}(MB2)).
Thus, we achieve a $4\%$ relative improvement of our DF stack, which retains its exact tree structure, by mapping to a deep ConvNet, training all weights by back-propagation, and mapping back to the original DF stack with updated threshold and leaf distributions.
However, note that the performance of the final DF is lower than the ConvNet, due to the approximate nature of the mapping.

\begin{figure*}[htp!]
\begin{center}
\begin{tabular}{cc}
\centering
   \includegraphics[height=0.215\textwidth]
                   {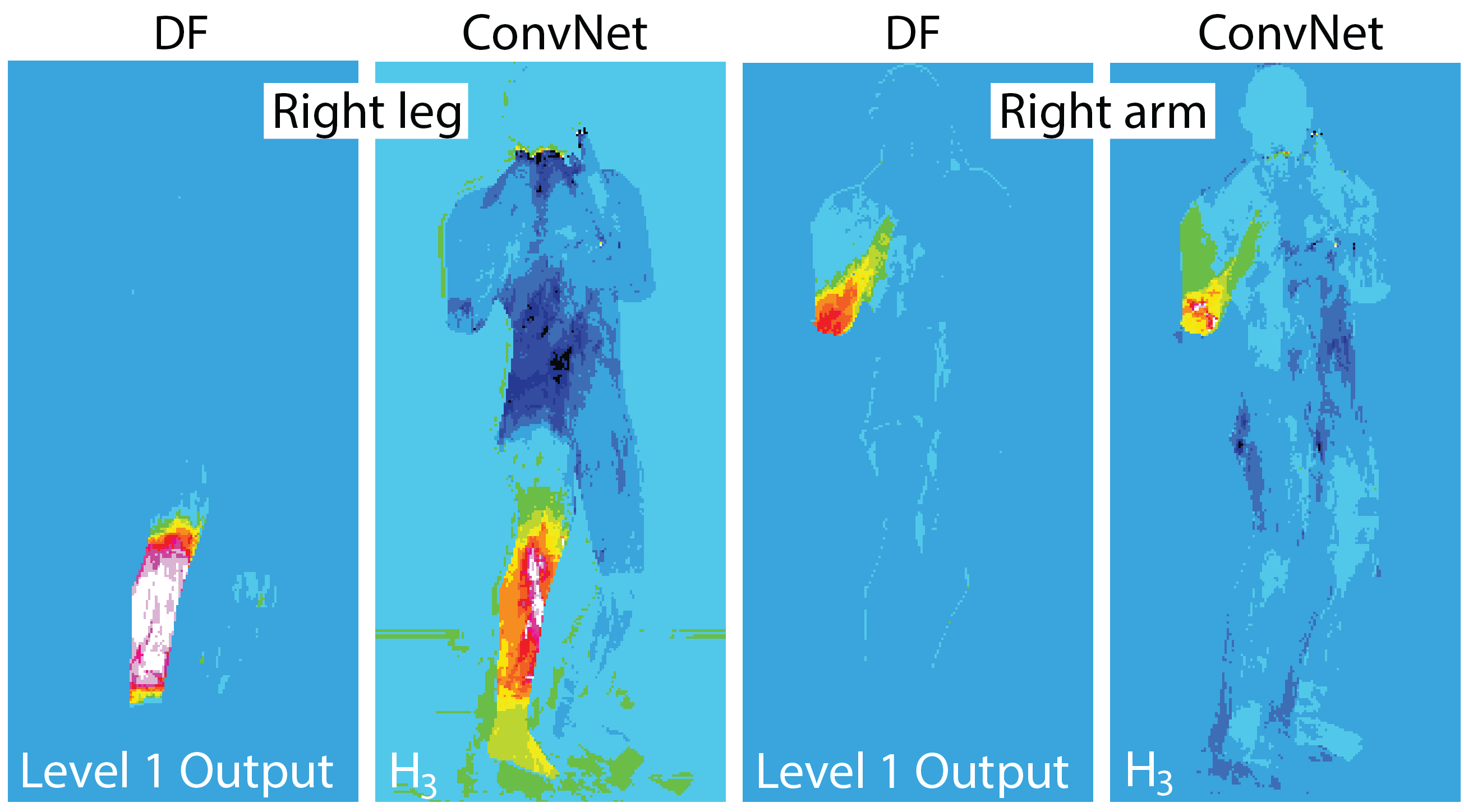} &
   \includegraphics[height=0.215\textwidth]
                   {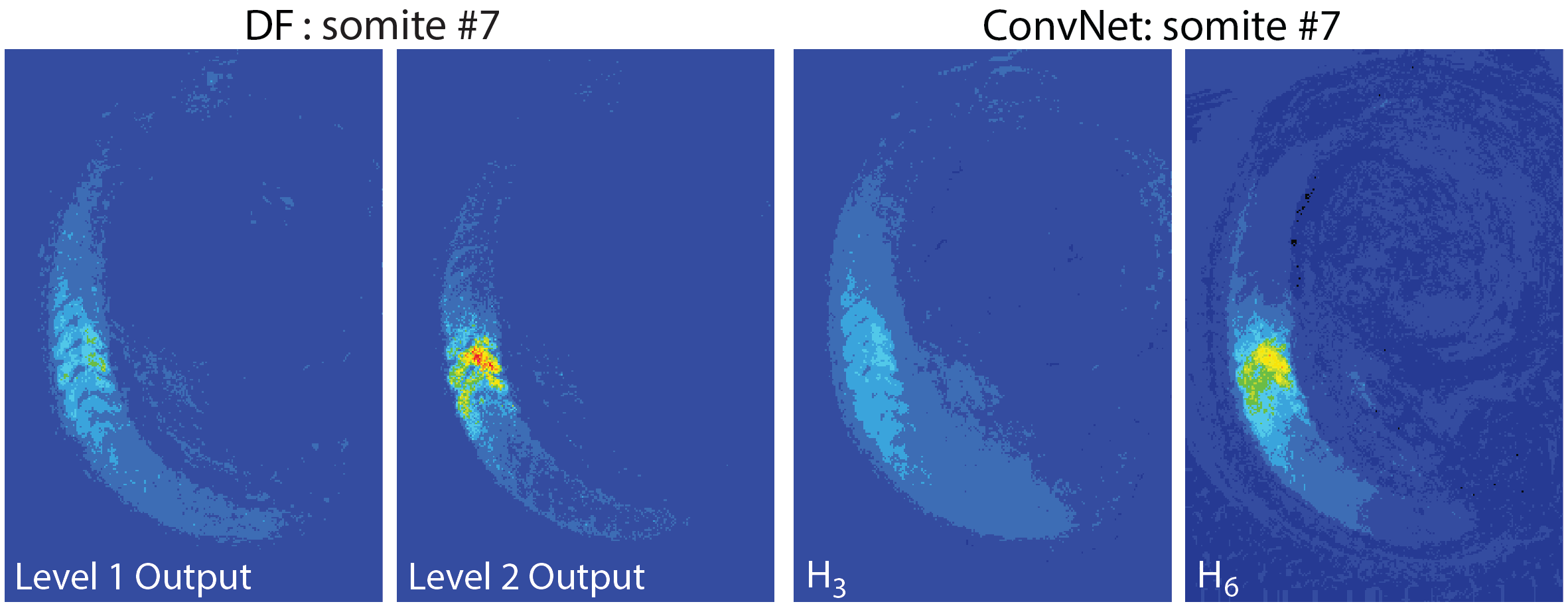} \\
	(a) & (b)
\end{tabular}
\end{center}
   \caption{\textbf{Visualization of internal activation layers.} We visualize the probability maps output by the intermediate output layers of the DF stack (\emph{e.g.,} Level 1,2 Output), and the activation maps from the corresponding hidden layers of the ConvNet (\emph{e.g.,} $H_3$, $H_6$) for (a) Kinect body parts, and (b) zebrafish somites. Notice that the activation from the ConvNet appears smoothed along the direction of the foreground classes compared to the noisier output of the stacked DF. Best viewed in colour}
\label{fig:activationLayers}
\end{figure*}

\textbf{Insights. }
The architecture of the deep ConvNet preserves the intermediate prediction layers of the DF stack, which generates one image for each class at the same resolution as the input image.
This enables us to gain insights on internal ConvNet layers.
However, due to back-propagation training, these images no longer represent probability distributions. In particular, the pixel values can now be negative. We visualized the internal layers to better understand how they changed during additional training in the ConvNet (Figure~\ref{fig:activationLayers}(a)).
Interestingly, we noticed that compared to the stacked DF, the internal activation layers in the ConvNet were less thresholded, and fired on adjacent body parts.
A common strategy in stacked classification is to introduce smoothing between the layers of the stack (\emph{e.g.}\ \cite{KontschiederKSC13, jampani, richmueller}), and it appears that a similar strategy is naturally learned by the deep ConvNet.

\subsection{Zebrafish Somite Classification}
\textbf{Experimental Setup. }
We next applied our method to semantic segmentation of 21 somites\footnote{Somites are the metameric units that give rise to muscle and bone, including vertebrae.} and 1 background class in a data set of 32 images (800x950 pixels) of developing zebrafish.
Experts in biology manually created ground truth segmentations of these images.
The data set was split into 16 images for training and 16 images for test.
Two additional training images were generated from each original training image by random rotation of the originals.
We evaluated the resulting segmentation by means of the class-balanced Dice score.

\textbf{Training of Stacked DF. }
We trained a three-level DF stack, with the following forest parameters at every level: 16 trees, maximum depth 12, stop node splitting if less than 25 samples. Features were extracted from the images using a standard filter bank, and then normalized to zero mean, unit variance. The number of random features tested in each node was set to the square root of the total number of input features. For each randomly selected feature, 10 additional contextual features were also considered, with X and Y offsets within a 129x129 pixel window. 
Training samples were generated by sub-sampling the training images 3x in each dimension and then randomly selecting $25\%$ of these samples for training.

\textbf{Training of ConvNet. }
We mapped the DF stack to a deep ConvNet with 8 hidden layers.
The ConvNet was initialized and trained exactly as for the Kinect example, with the following exeptions:
(i) We used a class-balanced cross-entropy loss function,
(ii) Training samples were generated by sub-sampling the training images 9x in each dimension.
(iii) Learning rate parameters were as follows: $a = 0.01$ and $b = 96$ iterations.
(iv) Momentum was initialized to $\mu = 0.4$, and increased to $0.7$ after $96$ iterations.
We observed convergence after only 1-2 passes through the training data, similar to what was reported by \cite{GirshickDDM14}.

\textbf{Training ConvNet from Random Initialization. }
For comparison to the DF-initialized weights described above, we also trained ConvNets with the same architecture, but with random weight initialization.
Weights were initialized according to a Gaussian distribution with zero mean and standard deviation, $\sigma = 0.01$.
We applied a similar SGD training routine, and re-tuned the hyper-parameters as follows: $a = 3$x$10^{-5}$, $b=96$ iterations, momentum was initialized to 0.4 and increased to 0.99 after 96 iterations.
Larger step-sizes failed to train.
Networks were trained for 2500 iterations.

\textbf{Training Parameters of Fully Convolutional Network. }
We also compared our method with the Fully Convolutional Network (FCN)~\cite{long_shelhamer_fcn_2015}.
This network was downloaded from Caffe's Model Zoo\footnote{\url{https://github.com/BVLC/caffe/wiki/Model-Zoo\#fcn}}, and initialized with weights fine-tuned from the ILSVRC-trained VGG-16 model.
We trained all layers of the network using SGD with a learning rate of $10^{-9}$, momentum of $0.99$ and weight decay of $0.0005$.

\begin{figure*}
\begin{center}
\begin{tabular}{cccccc}

   \includegraphics[height=0.21\textwidth]
                   {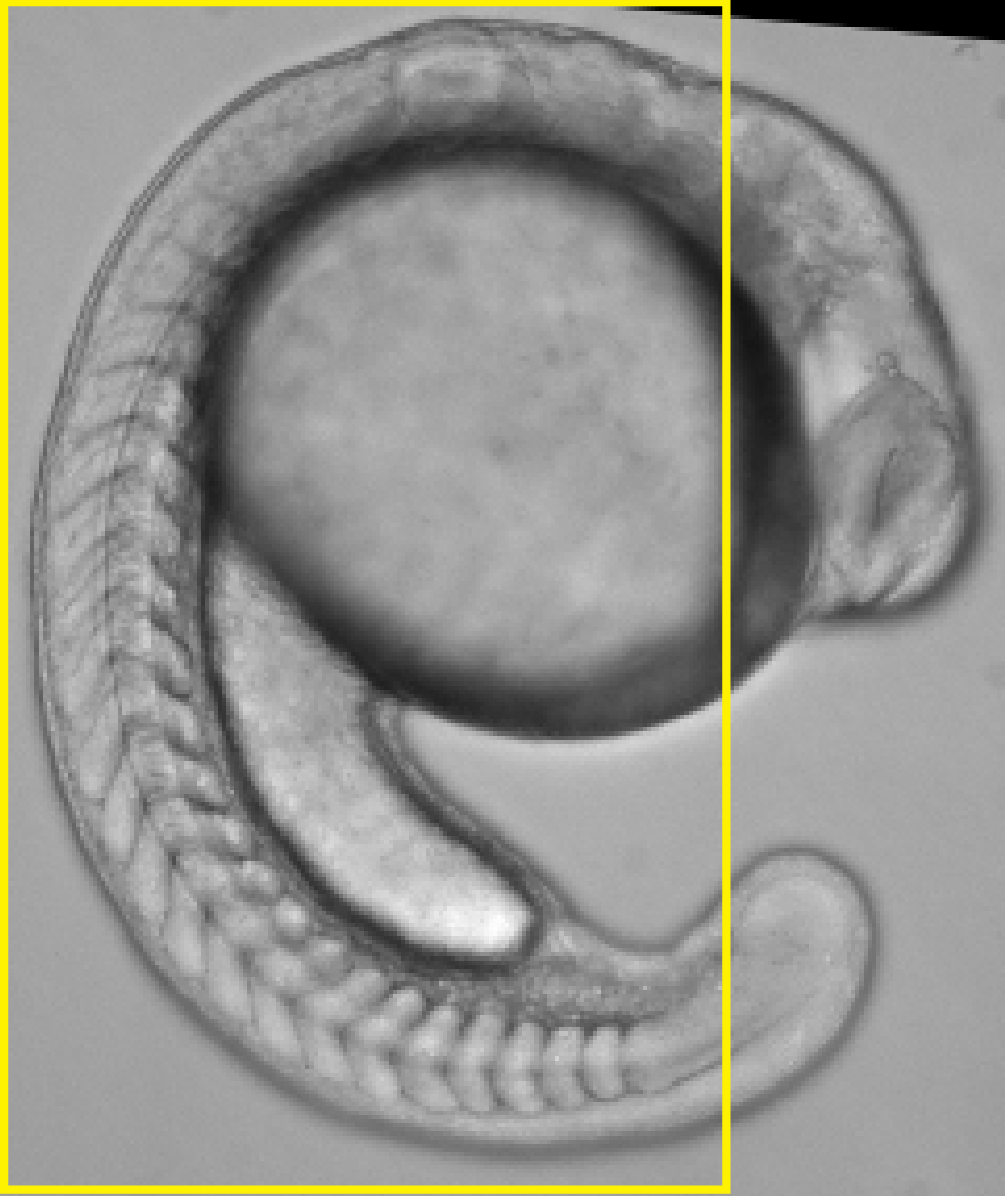} &
   \includegraphics[height=0.21\textwidth]
                   {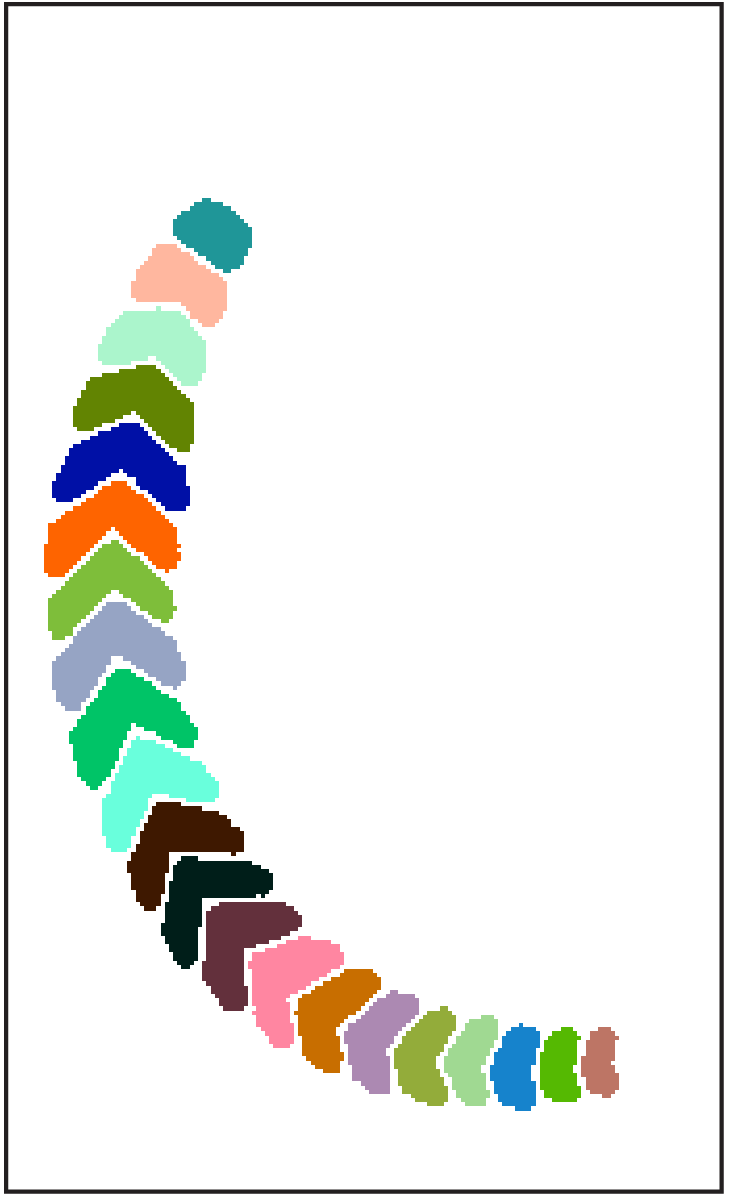} &
   \includegraphics[height=0.21\textwidth]
                   {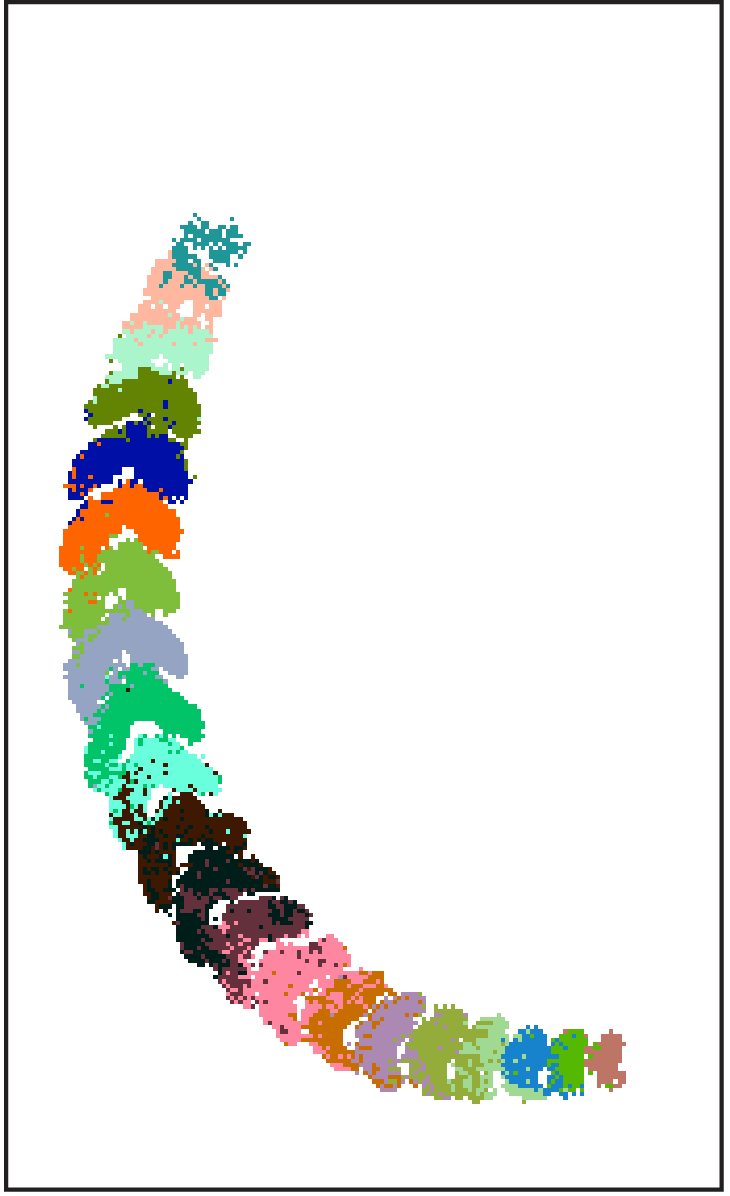} &
	 \includegraphics[height=0.21\textwidth]
                   {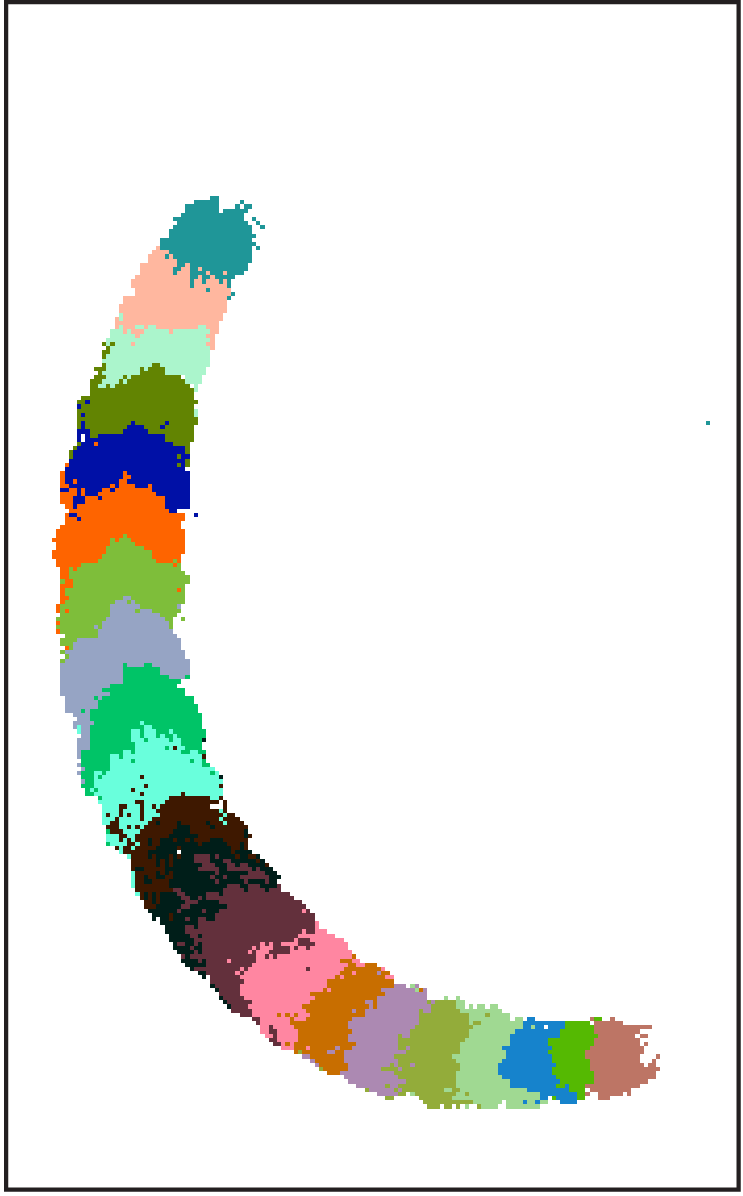} &
	 \includegraphics[height=0.21\textwidth]
                   {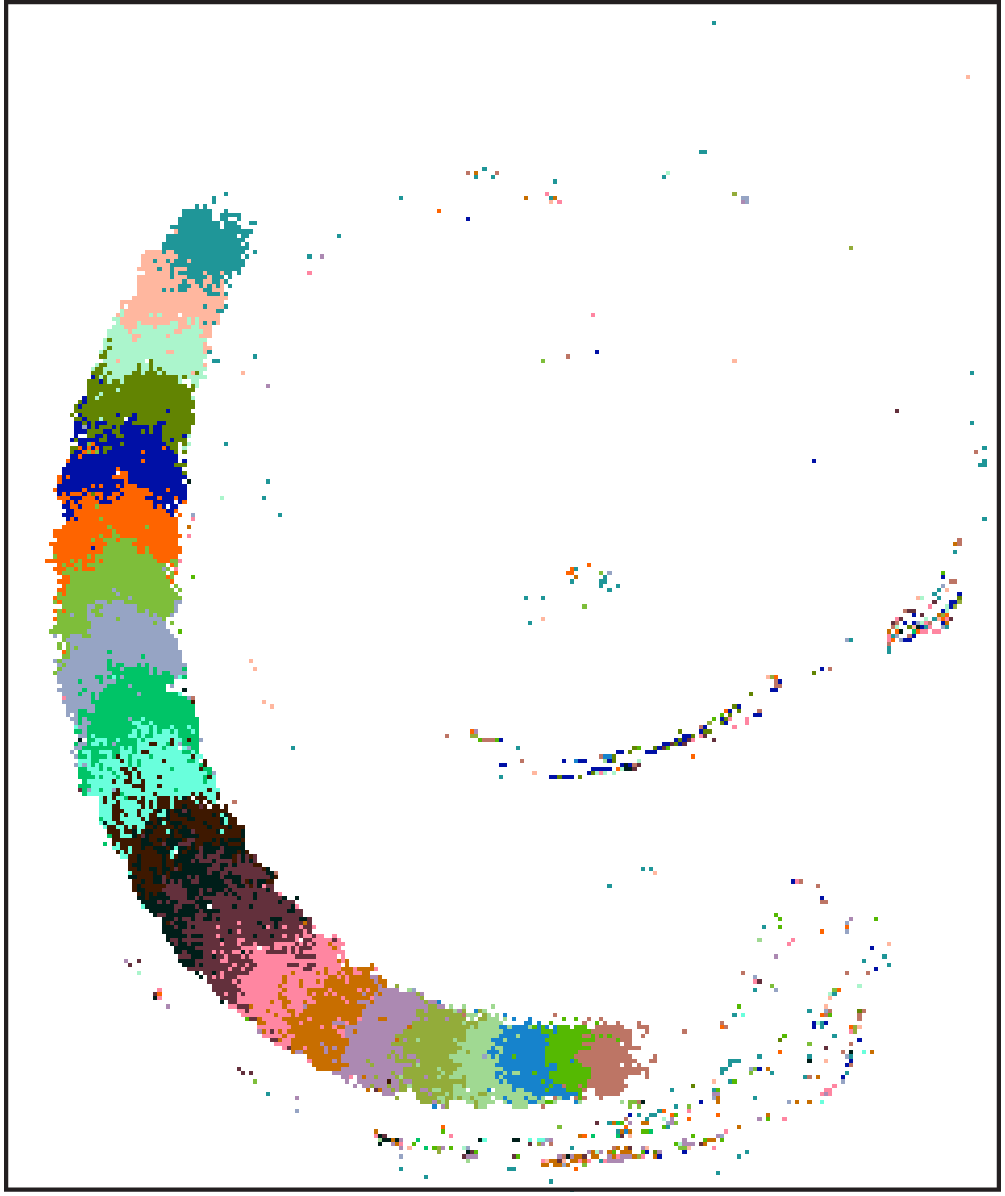} &
	 \includegraphics[height=0.21\textwidth]
                   {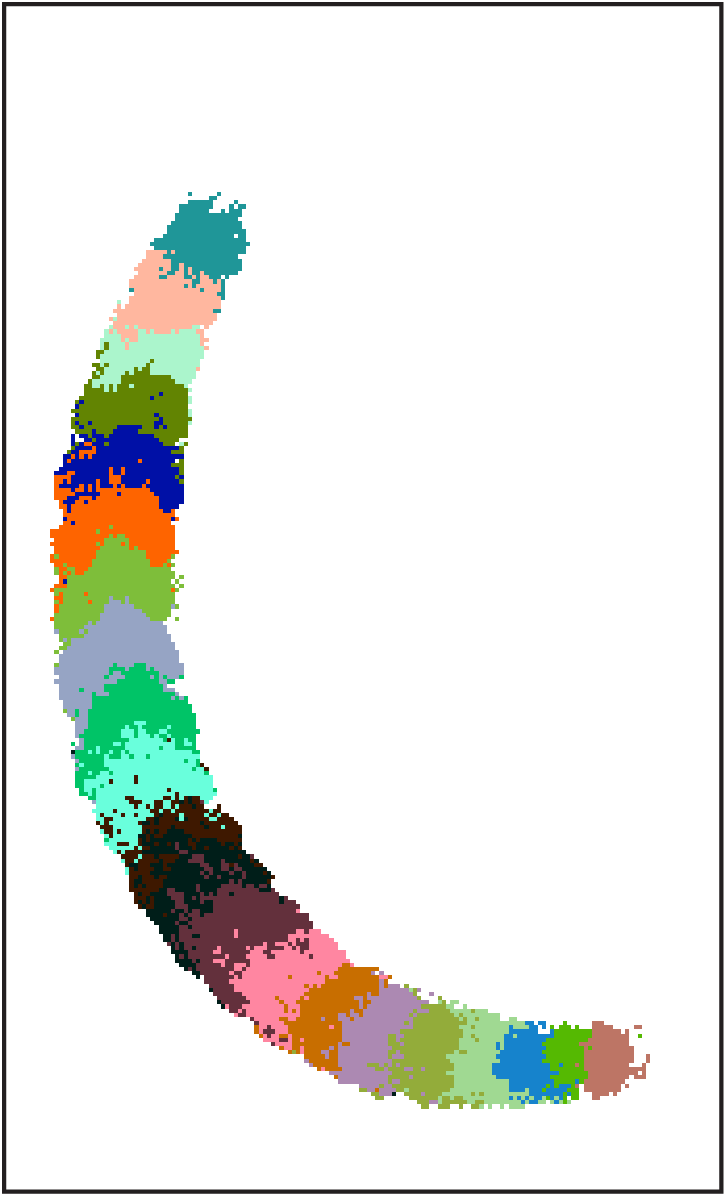} \\
	 			  (a) & (b) & (c) & (d) & (e) & (f)
\end{tabular}
\end{center}
   \caption{\textbf{Comparison of different methods for zebrafish somite labeling.} (a) Raw image of zebrafish.  Yellow box denotes crop for b,c,d,f. (b) Ground truth labeling. (c) Prediction of stacked DF. (d) Prediction of corresponding deep ConvNet, after parameter refinement by back-propagation. (e) Prediction of ``Map Back 1'' stacked DF. (f) Prediction of ``Map Back 2'' stacked DF. See Section~\ref{subsec:back-map} for details of map back algorithms.}
\label{fig:comparison}
\end{figure*}

\textbf{Results. }
Segmentation of the test data by means of the resulting three-level stacked DF achieved an average Dice score of 0.60
(see Figure~\ref{fig:comparison}(c)
and Table~\ref{tab:dice-score}(DF)).
The DF-initialized ConvNet achieved a Dice score of 0.66 after re-training, corresponding to a $10\%$ relative improvement
(see Figure~\ref{fig:comparison}(d)
and Table~\ref{tab:dice-score}(ConvNet)).
This result matches previous State-of-the-Art results on this data set \cite{richmueller}, but without the need for time-consuming model-based inference.
It's interesting to note that training a deep ConvNet with 8 hidden layers using hyperbolic tangent activation functions, and without batch normalization~\cite{batchnorm}, is typically extremely difficult, but works well here, likely due to the good initialization of the network.
We discuss insights on the internal activation layers of this network in Figure~\ref{fig:activationLayers}(b).
We also mapped the ConvNet back to the initial stacked DF architecture with updated parameters.
MB1 yielded a Dice score of $0.59$, and MB2 a final score of $0.63$, a $5\%$ improvement from the original DF model (Table~\ref{tab:dice-score} and Figure~\ref{fig:comparison}).

\begin{table}
  	\caption{\textbf{Comparison of dense semantic labeling. } Dice score is reported for the initial stacked DF (DF), DF-initialized and re-trained ConvNet (ConvNet), and after mapping the ConvNet back to a stacked DF using Map Back~1 and 2 (MB1 and MB2, respectively. See Section~\ref{subsec:back-map} for details). Higher Dice score corresponds to a more accurate segmentation.}
	\label{tab:dice-score}
	\begin{center}
		\begin{tabular}{lllll}
			\hline\noalign{\smallskip}
			& DF & ConvNet & MB1 & MB2  \\
			\noalign{\smallskip}\hline\noalign{\smallskip}
			\textbf{Kinect} & 0.82 & \textbf{0.91} & 0.74 & 0.85 \\
			\textbf{Zebrafish} & 0.60 & \textbf{0.66} & 0.59 & 0.63 \\
			\noalign{\smallskip}\hline
   		\end{tabular}
	\end{center}
\end{table}

We also considered the task of training the same ConvNet architecture from a random initialization.
We trained the network first maintaining the sparsity of the weight layers, and then fully connecting the layers corresponding to tree connectivity; however, these yielded final Dice scores of only $0.04$ and $0.15$, respectively.
Finally, we compared our method with the Fully Convolutional Network (FCN), a state-of-the-art model for semantic segmentation \cite{long_shelhamer_fcn_2015}.
This model (and others in the Caffe Zoo) take as input an RGB image, and are not directly amenable to grayscale microscopy images. We created 3-channel images by duplicating the grayscale image, and fine-tuned the network for approximately $1$ day on a single Nvidia K-40 GPU.
The FCN network failed to train successfully, achieving a Dice score of only $0.18$, due either to incompatibility of this pre-trained model with 1-channel images, the significant difference in task domain, and/or the limited size of the training set.

\section{Forward Mapping Revisited}
\label{sec:newmapping}

In Section~\ref{sec:method}, we demonstrated mapping a stacked DF to a deep ConvNet, based on the previously described mapping of a DT to a shallow NN~\cite{Sethi1990}.
These mappings are both exact when using DTs with thresholded split decisions; however, they yield NNs with non-differenti\-able activation functions, that are incompatible with training by back-propagation.
The compromise has been to model the split decisions with sharp, but differentiable activation functions, and then relax the network before training~\cite{Welbl14}.
This approach leads to two limitations:
(i) We observed that the initial performance of the NN, after relaxing but before training, decreased by as much as 10\% compared to the DF~\cite{Richmond:tc}.  We were able to overcome this loss during training of the NN; however, it's clear that the process of relaxation makes the initialization of the NN sub-optimal.
(ii) After training, the NN can no longer be mapped exactly back to a DF with thresholded split decisions.  We explored an approximate mapping; however, performance was degraded relative to the NN.

In the following we propose a new mapping that addresses these limitations: Instead of a DT with threshold split decisions, we map a fuzzy or soft DT with sigmoidal split decision~\cite{SuarezL99} to a NN.
A similar approach was taken by~\cite{Sethi:1995dt}; however, they used a heuristic ``winner takes all'' strategy, to preserve the $1$-of-$l$ encoding in the leaf layer.  Here, we derive a full probabilistic model of leaf node membership.
Surprisingly, our derivation leads to new insights into the recently proposed Concatenated Rectified Linear Unit (CReLU)~\cite{Shang:2016tj}, motivating it from a probabilistic perspective.

In a fuzzy DT~\cite{SuarezL99}, 
sigmoidal split functions reflect the probability of a sample going left vs. right at each split node. %
The choice of sigmoid function to model probabilistic split decisions can be understood by analogy to logistic regression with two categorical variables, in this case: left and right.
The probability of a sample belonging to leaf $l$, $p(\mathbf{x} \in X_l)$, can be expressed as the product of probabilities of going the correct direction at every split node along the path to this leaf.
\begin{multline}
	p(\mathbf{x} \in X_l) = \prod_{n \in P(l)} 
	[\ \textnormal{1}(X_l \subseteq X_{cl(n)})p(\mathbf{x} \in X_{cl(n)}) \\
	+ \textnormal{1}(X_l \subseteq X_{cr(n)})p(\mathbf{x} \in X_{cr(n)})\ ]
\label{eq:productofprobs}
\end{multline}
To map a fuzzy DT to a NN, we propose to employ two neurons per split node in Hidden Layer~1, such that one neuron, denoted $H^R_1(n)$, indicates the probability of the sample going right at split node $n$, $p(\mathbf{x} \in X_{cr(n)})$,
and the second neuron, $H^L_1(n)$, indicates the probability of the sample going left,
$p(\mathbf{x} \in X_{cl(n)})$. 
$H^L_1$ and $H^R_1$ are both connected to the input layer analogous to the original mapping described in Section~\ref{sec:method}, with identical weights and biases, $w_{{f(n)},H_1(n)} = \alpha_1$ and $b_{H_1(n)} = -\alpha_1 \cdot \theta_n$. As before we denote the resulting input as $z_{H_1(n)}$. 

In order to tackle the product in Equation~\ref{eq:productofprobs} via summation in Hidden Layer~2 of our proposed NN, we define the output of neurons in Hidden Layer~1 to be the $log$ of the sigmoid membership functions: 
$a_{H^R_1(n)}=log(\sigma(z_{H_1(n)}))$, and $a_{H^L_1(n)}=log(1-\sigma(z_{H_1(n)}))$.
We then let neurons in Hidden Layer 2 calculate the probability of leaf membership in Equation~\ref{eq:productofprobs} by computing the sum of log likelihoods, and then exponentiating the result to undo the $log$. 
The weights encode the indicator functions in equation~\ref{eq:productofprobs}: $w_{H^R_1(n),H_2(l)} = 1$ if $X_l \subseteq X_{cr(n)}$, and $w_{H^L_1(n),H_2(l)} = 1$ if $X_l \subseteq X_{cl(n)}$.
All other weights and biases are zero.

\begin{figure*}[htp!]
\begin{center}
\begin{tabular}{cc}

   \includegraphics[height=0.25\textwidth]
                   {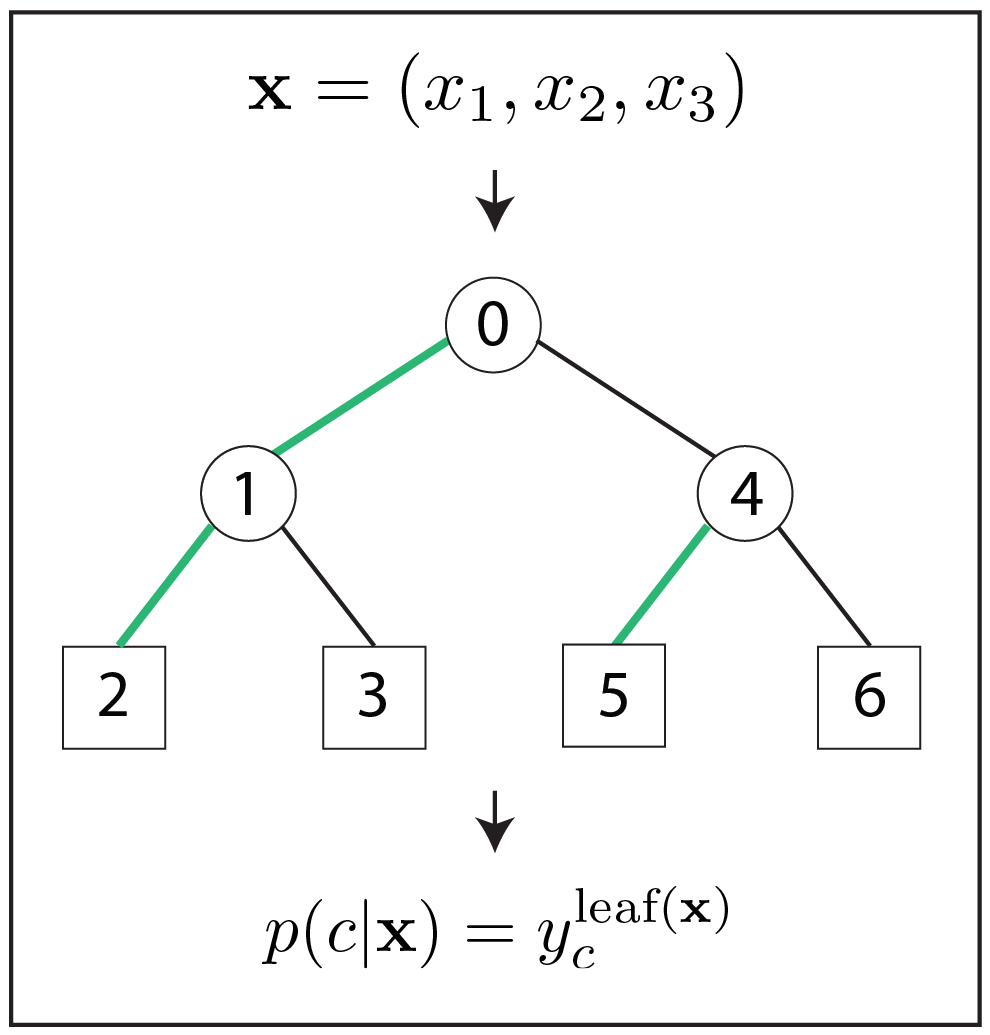} &
   \includegraphics[height=0.25\textwidth]
                   {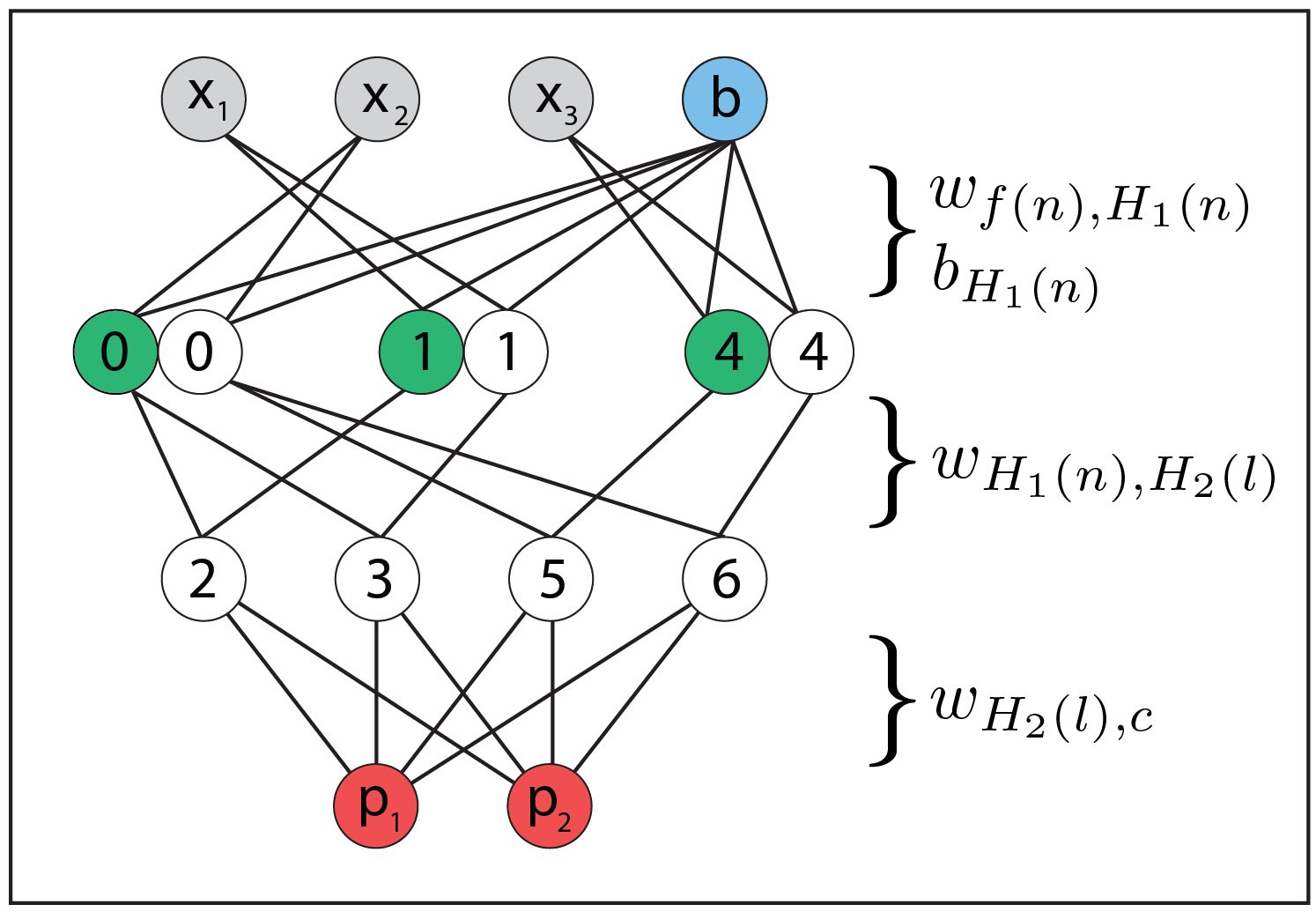} \\
				   (a) & (b)
\end{tabular}
\end{center}
   \caption{\textbf{Mapping from a fuzzy DT to a NN.} (a) A shallow DT. The architecture of the DT is equivalent to Figure~\ref{fig:forward-map_TREE}(a) to highlight the difference between mapping DTs with threshold vs soft split decisions.  (b) Corresponding NN with two hidden layers.  Contrary to the previous mapping, the first hidden layer contains two neurons for every split node, $H^L_1(n)$ (green), and $H^R_1(n)$ (white).  Note that leftward paths in the tree, corresponding to $H^L_1(n)$, are also highlighted in green.  Both neurons share the same connections and weights, $w_{f(n),H_1(n)}$; however, they have different activation functions.  $H^L_1(n)$ has activation function $-\textnormal{RELU}(z_{H_1(n)})$, corresponding to the log probability of the sample going left at this node.  $H^R_1(n)$ has activation function $-\textnormal{RELU}(-z_{H_1(n)})$, corresponding to the log probability of the sample going right at this node.  
The weights $w_{H_1(n),H_2(l)}$ between the two hidden layers encode the structure of the tree.
For example, leaf 5 is connected to split node $H^R_1(n_0)$ and $H^L_1(n_4)$ because $X_{l_5} \subseteq X_{cl(n_4)} \subseteq X_{cr(n_0)}$.
The final weights $w_{H_2(l),c}$ are fully connected and store the votes $y_c^l$ for each leaf $l$ and class $c$. Same color coding and node labeling as in Figure~\ref{fig:forward-map_TREE}}
\label{fig:fuzzyDTmapping}
\end{figure*}

The output layer is unchanged from the original mapping, and computes the weighted average of the leaf distributions, with the weight corresponding to the probability of leaf membership.  
See Figure~\ref{fig:fuzzyDTmapping} for the network architecture.
Note that while an exponential activation function in $H_2$ would undo the effect of the log transformation in $H_1$, it may be advisable to use a softmax activation in this layer, to ensure a normalized probability distribution over leafs.

Our proposed mapping is closely related to the original mapping of a conventional DT with threshold split decisions, but importantly is interpretable as a fuzzy or soft DT, even after relaxation and further training of the network.  

\begin{figure}[htp!]
\begin{center}
\begin{tabular}{cc}

   \includegraphics[width=0.23\textwidth]
                   {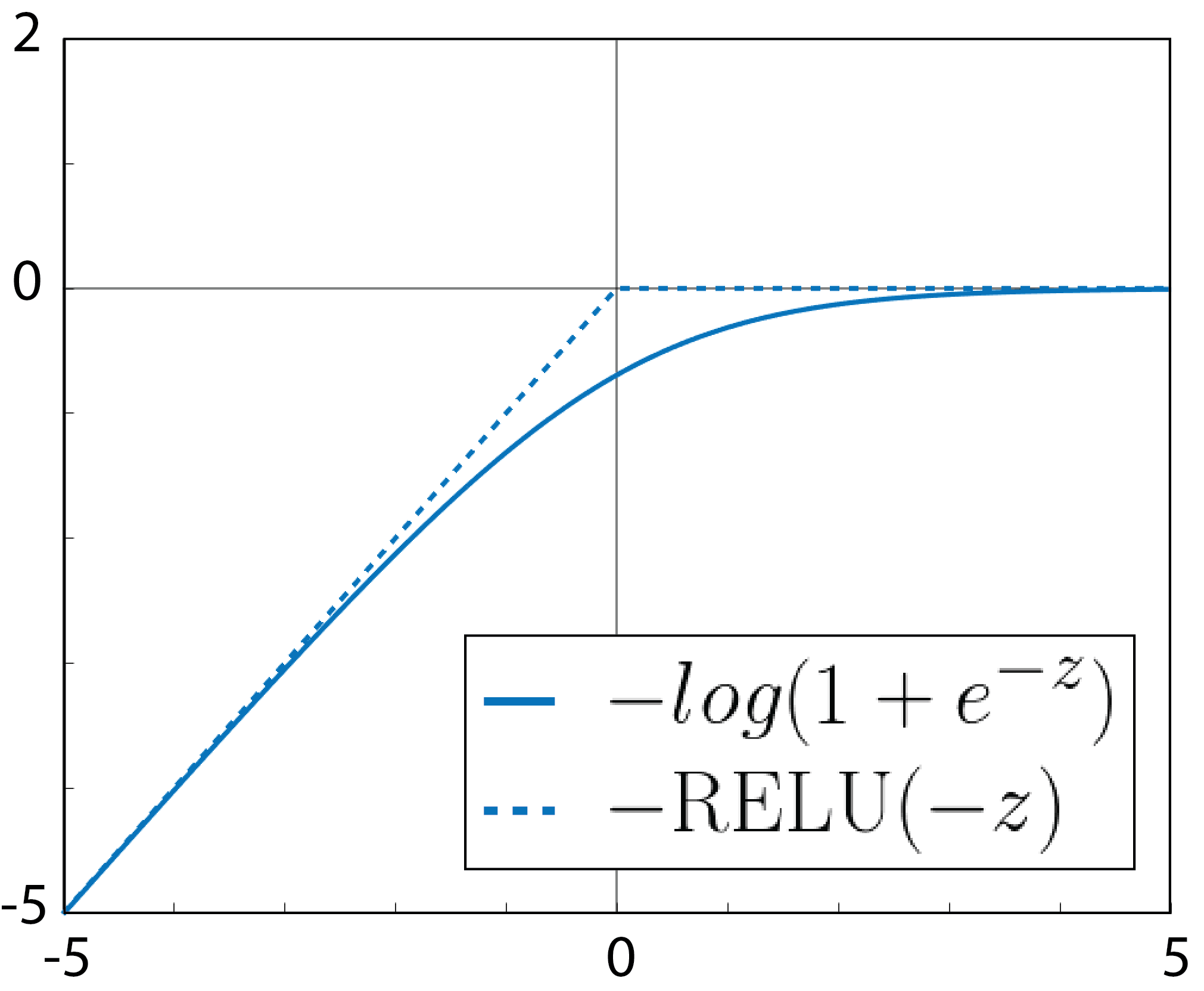} &
   \includegraphics[width=0.23\textwidth]
                   {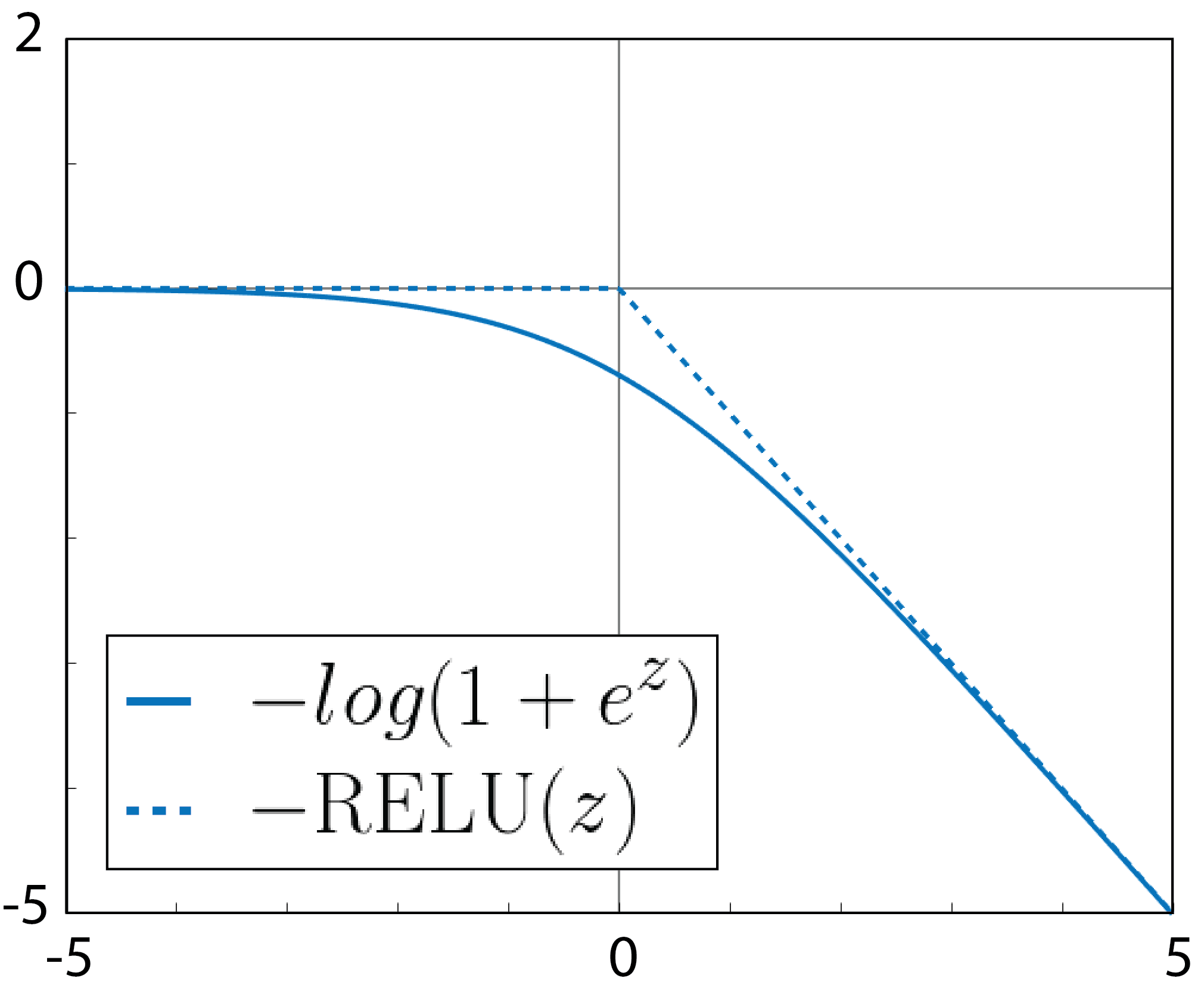} \\
				   (a) & (b)
\end{tabular}
\end{center}
   \caption{\textbf{Proposed activation functions for Hidden Layer 1.} (a) The exact form for $a_{H^R_1(n)}$ is shown (solid blue line), as well as its approximation using $\textnormal{ReLU}$ (dotted blue line).  (b) Similarly for $a_{H^L_1(n)}$}
\label{fig:ReLU}
\end{figure}

Furthermore, this mapping provides new insight for the role of ReLU~\cite{Nair:2010vq, Glorot:2011tm} in NNs as a transformation from probability to log probability space. To this end, note the following relationship: 
\begin{equation}
\begin{split}
	a_{H^R_1(n)} 
	& = log(p(\mathbf{x} \in X_{cr(n)})) \\
	& = log(\sigma(z_{H_1(n)})) \\
	& = - log(1+e^{(-z_{H_1(n)})}) \\
	& \approx min(0,z_{H_1(n)}) \\ 
	& = -\textnormal{ReLU}(-z_{H_1(n)}).
\end{split}
\end{equation}
Analogously, \mbox{$a_{H^L_1(n)} \approx -\textnormal{ReLU}(z_{H_1(n)})$}.
See Figure~\ref{fig:ReLU} for visualizations of these activation functions.
Hence our mapping motivates ReLU from the perspective of fuzzy DT models, where it can be seen as computing the log probability associated with membership to a region in feature space.

Interestingly, the paired features that arise in hidden layer 1 of our model, $(a_{H^R_1(n)},a_{H^L_1(n)})$, correspond to the use of concatenated ReLU $([z]_{+},\allowbreak[-z]_{+})$ in~\cite{Shang:2016tj}, except for a change of sign, which can simply be absorbed into the weights.
This activation function was recently proposed based on the empirical observation that \textit{e.g.}, AlexNet~\cite{krizhevsky_cnn_2012} learns pairs of strongly negatively correlated kernels in its first convolutional layer~\cite{Shang:2016tj}.  It has also been observed by numerous other groups that explicitly constraining a ConvNet to produce such paired or concatenated ReLU kernels leads to increased performance~\cite{Blot:2016ie, Coates:2011ud,Kim:2015us}.

Our findings are in a similar vein to~\cite{Patel}, which derives a first principles explanation of common operations, including ReLU, found in deep ConvNets, from the perspective of max-sum inference in generative Deep Rendering Mixture Models.
By contrast, starting from a soft DT model, we demonstrate that concatenated ReLU can be directly interpreted as an approximation to the log probability of a sample belonging to the left and right half-spaces associated with a decision boundary.

\section{Conclusions and Future Work}

We have exploited a new mapping between stacked DFs and deep ConvNets, and demonstrated the benefits for semantic segmentation with limited training data.
Furthermore, we generalized this mapping to include fuzzy DTs with soft split decisions, and in the process garnered new insights into the CReLU activation function.

There are many exciting avenues for future research.
First, it will be interesting to compare the performance of the original mapping against the new mapping with soft split decision.
In particular, the use of ReLU activation functions in the new mapping may enable deeper models to be trained more efficiently, as has been widely observed in other ConvNet models.
We would also like to compare the performance of our model directly to the recent model from \cite{DeepNDFs}, and explore whether our Auto-context ConvNet can be incorporated as a module after the feature extraction layers in a traditional ConvNet.

Finally, there are interesting possibilities for exploiting the fact that each hidden layer in our proposed ConvNet is directly interpretable.
For example, our ConvNet architecture produces internal activation images that correspond to class probabilities.
Thus, it would be straightforward to incorporate differentiable model layers, \emph{e.g.,}~\cite{Girshick:2014um}, that operate on the output of these layers.
There are also interesting opportunities for applying structured regularization.  For example, $L_1$ regularization on activations in the leaf layers (\textit{e.g.}, $a_{H_2(l)}$) would encourage samples to traverse only a small number of paths in the complementary fuzzy tree.

\bibliographystyle{spmpsci}      
\bibliography{Richmond_Kainmueller.bib}   

\end{document}